%% file: _main.tex
\input{_constants}
\arxiv 

\pdfoutput=1
\documentclass[10pt,twocolumn,letterpaper]{article}
\input{cvpr_header}
\begin{document}
\title{\paperTitle}
\author{\authorBlock}
\maketitle

\input{00_abstract}
\input{01_intro}
\input{02_related}

\input{03_method}

\input{04_experiments}

\input{10_conclusion}

\section*{Appendix}

This appendix encompasses comprehensive attribute results from the LaSOT~\cite{fan2019lasot} datasets. Additionally, it delineates the performance achieved on the TNL2K~\cite{wang2021tnl}, NFS~\cite{kiani2017need}, and UAV123~\cite{mueller2016uav} datasets. Subsequent sections detail extensive ablation experiments, meticulously examining the impacts of sequence format, trajectory-appearance evolution, and the masking ratio within the appearance model. Finally, it includes an array of rich visualizations, encompassing additional cross-attention maps and image reconstruction visualizations.

\input{12_appendix}

{\small
\bibliographystyle{ieee_fullname}
\bibliography{11_references}
}

\end{document}

%% file: _constants.tex
\def\cvprPaperID{3386}
\def\confName{CVPR}
\def\confYear{2024}

\def\paperTitle{ARTrackV2: Prompting Autoregressive Tracker Where to Look and \\How to Describe}

\def\authorBlock{
    Yifan Bai\qquad
    Zeyang Zhao\qquad
    Yihong Gong\qquad
    Xing Wei\,\textsuperscript{\Letter}\\[2mm]

    Xi'an Jiaotong University \qquad \\
    {\tt\small 
    \{yfbai, zeyang\}@stu.xjtu.edu.cn \qquad 
    \{ygong, weixing\}@mail.xjtu.edu.cn }\\
    \url{https://ARTrackV2.github.io/}
}

\newif\ifreview 
\newif\ifarxiv \newcommand{\arxiv}{\arxivtrue}
\newif\ifcamera 
\newif\ifrebuttal 

%% file: cvpr_header.tex
\ifreview \usepackage[review]{cvpr} \fi
\ifarxiv \usepackage[pagenumbers]{cvpr} \fi
\ifrebuttal \usepackage[rebuttal]{cvpr} \fi
\ifcamera \usepackage{cvpr} \fi

\usepackage{graphicx}
\usepackage{amsmath}
\usepackage{amssymb}
\usepackage{booktabs}

\input{_macros}

\usepackage{xr-hyper}

\makeatletter
\newcommand*{\addFileDependency}[1]{
  \typeout{(#1)}
  \@addtofilelist{#1}
  \IfFileExists{#1}{}{\typeout{No file #1.}}
}

\makeatother

\usepackage[pagebackref,breaklinks,colorlinks]{hyperref}
\usepackage[capitalize]{cleveref}
\crefname{section}{Sec.}{Secs.}
\crefname{table}{Table}{Tables}
\crefname{figure}{Fig.}{Figs.}

%% file: _macros.tex
\usepackage{times}
\usepackage{microtype}
\usepackage{epsfig}
\usepackage[table,xcdraw]{xcolor}
\usepackage{pifont}
\usepackage{caption}
\usepackage{float}
\usepackage{placeins}
\usepackage{color, colortbl}
\usepackage{stfloats}
\usepackage{enumitem}
\usepackage{tabularx}
\usepackage{xstring}
\usepackage{multirow}
\usepackage{xspace}
\usepackage{url}
\usepackage{marvosym}
\usepackage{subcaption}
\usepackage{xcolor}
\usepackage[hang,flushmargin]{footmisc}
\usepackage{nicematrix}
\usepackage{amsmath,amsfonts,bm}
\usepackage{newpxmath}
\usepackage{tikz}
\usepackage{pgfplots}
\usepackage[export]{adjustbox} 
\usepackage{graphicx}
\usetikzlibrary{plotmarks,patterns}
\pgfplotsset{compat=newest}
\usepackage{ulem}

\definecolor{cbins}{RGB}{40, 120, 181}
\definecolor{cexbins}{RGB}{196, 49, 25}
\definecolor{cexexbins}{RGB}{246, 189, 96}
\ifcamera \usepackage[accsupp]{axessibility} \fi

% ...

\usepackage{graphicx, amsmath, amssymb, caption, subcaption, multirow, overpic, textpos}
\usepackage[table]{xcolor}
\usepackage[british, english, american]{babel}
\definecolor{citecolor}{HTML}{0071BC}
\definecolor{linkcolor}{HTML}{ED1C24}

\newlength\savewidth
\renewcommand{\paragraph}[1]{\vspace{1.25mm}\noindent\textbf{#1}}

\newcolumntype{x}[1]{>{\centering\arraybackslash}p{#1pt}}
\newcolumntype{y}[1]{>{\raggedright\arraybackslash}p{#1pt}}
\newcolumntype{z}[1]{>{\raggedleft\arraybackslash}p{#1pt}}

\newcommand{\app}{\raise.17ex\hbox{$\scriptstyle\sim$}}

\definecolor{deemph}{gray}{0.6}

\definecolor{baselinecolor}{gray}{.9}

\newcommand{\R}[1]{{%
    \textbf{%
        \ifstrequal{#1}{1}{\textcolor{red}{R#1}}{%
        \ifstrequal{#1}{2}{\textcolor{blue}{R#1}}{%
        \ifstrequal{#1}{3}{\textcolor{magenta}{R#1}}{%
        \ifstrequal{#1}{4}{\textcolor{teal}{R#1}}{%
                           \textcolor{cyan}{R#1}%
        }}}}%
    }%
}}

\definecolor{defaultcolor}{gray}{.9}
\newcommand{\default}[1]{\cellcolor{defaultcolor}{#1}}

\definecolor{cARTrack-L-384}{RGB}{250,155, 143}
\definecolor{cARTrack-384}{RGB}{40, 120, 181}
\definecolor{cARTrack-256}{RGB}{255, 210, 150}
\definecolor{cOSTrack}{RGB}{200,194, 230}
\definecolor{cMixFormer-L}{RGB}{153, 153, 153}
\definecolor{cSTARK}{RGB}{247, 237, 226}
\definecolor{cTransT}{RGB}{246, 178, 147}
\definecolor{cDiMP}{RGB}{220, 109, 87}
\definecolor{cATOM}{RGB}{241, 217, 203}
\definecolor{cOcean}{RGB}{255,175, 204}
\definecolor{cECO}{RGB}{130, 177, 211}
\definecolor{cSiamFC}{RGB}{246, 189, 96}

\definecolor{cSiamRCNN}{RGB}{123, 129, 164}
\definecolor{cSiamBiAtt}{RGB}{165, 174, 190}
\definecolor{cDiMP}{RGB}{166, 135, 132}
\definecolor{cCTTrack}{RGB}{180,210,251}
\definecolor{cGRM}{RGB}{255, 165, 180}
\definecolor{cMixViT}{RGB}{193, 193, 193}
\definecolor{cSeqTrack}{RGB}{247, 237, 226}
\definecolor{groundtruth}{RGB}{0, 176, 80}
\definecolor{cARTrackV2-L-384}{RGB}{250,127, 111}
\definecolor{cARTrackV2-384}{RGB}{40, 120, 181}
\definecolor{cARTrackV2-256}{RGB}{255, 190, 122}

\definecolor{cSiamRPN++}{RGB}{242, 159, 5}
\definecolor{cSiamMask}{RGB}{100, 80, 200}

\definecolor{cbins}{RGB}{40, 120, 181}
\definecolor{cexbins}{RGB}{196, 49, 25}
\definecolor{cexexbins}{RGB}{246, 189, 96}
\definecolor{seq}{RGB}{170, 196, 135}

\definecolor{mygreen}{RGB}{0,135,54}
\definecolor{myblue}{RGB}{33,113,181}
\definecolor{myred}{RGB}{201,0,31}
\definecolor{myyellow}{RGB}{255,255,0}
\definecolor{mypurple}{RGB}{92,58,150}
\definecolor{maskgreen}{RGB}{170,196,135}
\definecolor{mygray}{gray}{0.6}

\definecolor{defaultcolor}{gray}{.9}

%% file: 00_abstract.tex
\begin{abstract}
We present ARTrackV2, which integrates two pivotal aspects of tracking: determining \textbf{where to look} (localization) and \textbf{how to describe} (appearance analysis) the target object across video frames. Building on the foundation of its predecessor, ARTrackV2 extends the concept by introducing a unified generative framework to \textbf{``read out"} object's trajectory and \textbf{``retell"} its appearance in an autoregressive manner. This approach fosters a time-continuous methodology that models the joint evolution of motion and visual features, guided by previous estimates. Furthermore, ARTrackV2 stands out for its efficiency and simplicity, obviating the less efficient intra-frame autoregression and hand-tuned parameters for appearance updates. Despite its simplicity, ARTrackV2 achieves state-of-the-art performance on prevailing benchmark datasets while demonstrating remarkable efficiency improvement. In particular, ARTrackV2 achieves AO score of 79.5\% on GOT-10k, and AUC of 86.1\% on TrackingNet while being $3.6 \times$ faster than ARTrack. The code will be released.

\end{abstract}

%% file: 01_intro.tex
\section{Introduction}
\label{sec:intro}

\begin{figure}
    \centering
    \includegraphics[width=1.0\linewidth]{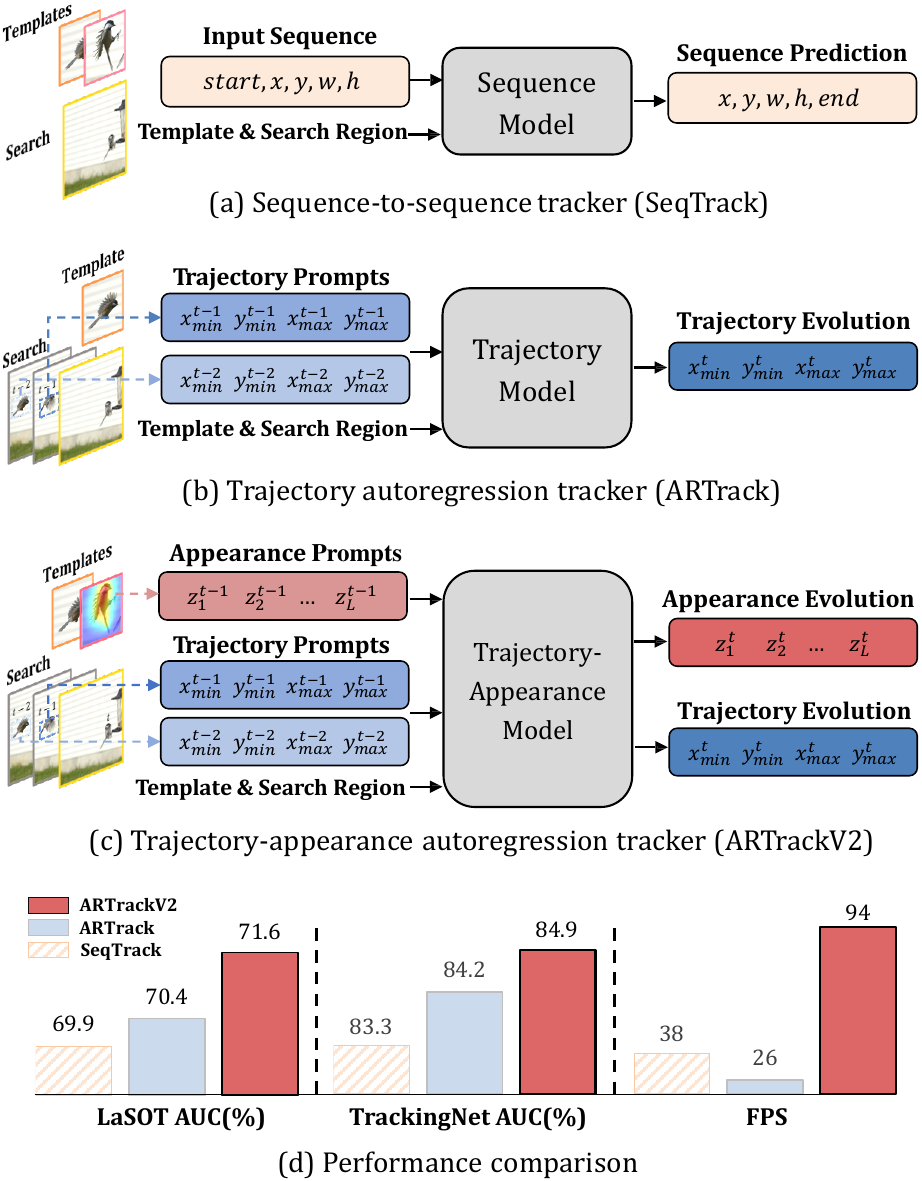}
    \caption{\textbf{Frameworks and performance comparison} of trackers following the sequence generation paradigm. (a) SeqTrack views tracking as sequence prediction. (b) ARTrack introduces trajectory evolution. (c) ARTrackV2 incorporates joint trajectory-appearance evolution. (d) Performance comparison.}
    \label{fig:compare}
    \vspace{-8pt}
\end{figure}

Visual object tracking~\cite{fan2019siamese,li2018learning,bhat2018unveiling,nam2016learning,kiani2017learning,javed2022visual}, a cornerstone in the realm of computer vision, has witnessed transformative advancements over the past decade. Its applications span a diverse array of fields, from autonomous vehicles to surveillance, and from augmented reality to human-computer interaction. At its core, visual tracking involves the continuous localization of an object within a video sequence, typically initiated from the first frame.

In this research area, previous approaches have primarily focused on either trajectory estimation or appearance modeling. Traditional methods, such as the application of Kalman filters~\cite{bertozzi2004pedestrian,weng2006video,chen2011kalman} and Particle filters~\cite{gustafsson2002particle, arulampalam2002tutorial}, emphasize predicting the object's motion by leveraging historical states. In contrast, modern learning-based methods strive to understand and track the visual features of the target object, often employing a template-matching framework.
However, these approaches typically adopt frame-level training strategies, overlooking the temporal dependencies across frames. Some methods attempt to handle appearance changes over time using dynamic templates, which are updated using heuristic rules~\cite{fu2021stmtrack} or learnable modules~\cite{stark, mixformer, cui2023mixformerv2}.

The recent shift towards a generative paradigm~\cite{chen2021pix2seq} in visual tracking~\cite{ARTrack,SeqTrack}, conceptualizing the task as sequence generation, has set new performance benchmarks. This approach simplifies the process, directly predicting object coordinates sequentially.
SeqTrack~\cite{SeqTrack}, as shown in Figure~\ref{fig:compare}(a), introduces an \textit{intra-frame} sequence model that generates four tokens of the bounding box autoregressively. 
It also showcases that prepending previous coordinate tokens in inference could improve accuracy further.
In contrast, ARTrack~\cite{ARTrack} concentrates on \textit{inter-frame} autoregression (Figure~\ref{fig:compare}(b)). 
It advocates for video sequence-level training (rather than frame-level) to maintain consistency between training and testing in terms of data distributions and task objectives~\cite{slt}.
Overall, this generative framework has its flexibility to utilize historical trajectory tokens, referred to as \textit{trajectory prompts}, to model trajectory evolution continually.

In this paper, we go one step further and introduce a \textit{joint trajectory-appearance} autoregression tracker. Building upon the foundation of its predecessor, ARTrackV2 extends the concept by implementing a unified generative framework that models the evolution of both trajectory and appearance. 
The intuition behind this idea is simple: \textbf{if the tracker successfully tracks an object, it should not only ``read out" object's position but also ``retell" its appearance}.
Alongside the time-series modeling of trajectory proposed by ARTrack, we maintain an autoregressive model to simultaneously reconstruct the object's appearance, using a set of \textit{appearance prompts}, as illustrated in Figure~\ref{fig:compare}(c).
These tokens, on the one hand, function similarly to dynamic templates, interacting with the search region through attention mechanisms. Beyond that, they are trained to rebuild the object's appearance, requiring an understanding of visual feature evolution. We design a masking strategy that intentionally prevents the attention from appearance tokens to trajectory ones, preventing the appearance model from merely cropping visual features based on the predicted trajectory.

Furthermore, ARTrackV2 distinguishes itself through its operational simplicity and efficiency. 
Different from SeqTrack and ARTrack, it utilizes a pure encoder architecture to process all tokens within a frame, in parallel. ARTrackV2 abandons intra-frame autoregression that impedes tracking efficiency while maintaining the time-autoregressive framework (aka, inter-frame autoregression). In contrast to many contemporary tracking systems that require multiple training stages~\cite{mixformer, cui2023mixformerv2, stark} or hand-tuned parameters for template updates~\cite{SeqTrack, fu2021stmtrack}, ARTrackV2 undergoes end-to-end training within a single stage. This approach yields outstanding performance on various benchmark datasets, with the base model achieving an impressive AUC score of 71.6\% on LaSOT and 84.9\% on TrackingNet. Notably, it accomplishes this while exhibiting a substantial $3.6\times$ speed improvement compared to ARTrack, as demonstrated in Figure~\ref{fig:compare}(d).
our top-performing model achieves an even higher AUC score of 73.6\% on LaSOT and an impressive 86.1\% on TrackingNet, significantly outperforming SeqTrack and ARTrack while maintaining remarkable speed improvements of approximately $3\times$ to $5\times$.

To summarize, ARTrackV2 enhances its predecessor in the following ways:
\begin{itemize}
    \item \textbf{Extend the concept}: we complement the generative framework for visual tracking to encompass both trajectory generation and appearance reconstruction.
    \vspace{-1mm}
    \item \textbf{Strengthen inter-frame autoregression}: we uphold the time-autoregressive model to jointly evolve trajectory and appearance. Also, we introduce sequence data augmentation to improve accuracy.
    \vspace{-1mm}
    \item \textbf{Eliminate intra-frame autoregression}: we employ a pure encoder architecture that enables parallel processing of all tokens within a frame, moving away from the less efficient intra-frame autoregressive decoder.
\end{itemize}

%% file: 02_related.tex
\section{Related Work}
\label{sec:related}

{\flushleft\textbf{Tracking Framework.}}\quad
Prevailing trackers~\cite{TransT, danelljan2019atom, li2019siamrpn++, siamFC, dai2020high, guo2020siamcar, zhu2018distractor} often employing a template-matching framework reference template to match target within the search region.
Initially, these approaches employ a backbone to integrate visual features\cite{ostrack, mixformer, chen2022backbone, swin}, then split tracking into multiple subtasks\cite{bhat2019learning, aiatrack, li2018high, song2022transformer, raffel2020exploring, Ocean} such as object scale estimation and center point localization, divide and conquer with specific heads. Meanwhile, they introduce complex post-processings, overlooking potential temporal dependencies.
Recently, generative paradigm~\cite{ARTrack, SeqTrack} redefines tracking as a sequence generation task. After visual integration, this approach unified multiple tracking subtasks as an intra-frame sequence model in an autoregressive manner. This methodology simplified the tracking framework and leveraged preceding trajectory tokens to model trajectory evolution but impeded efficiency by introducing intra-frame autoregression. Thus, we propose a pure encoder architecture that enables parallel processing of all tokens, abandoning intra-frame autoregression while preserving a time-autoregression nature.

{\flushleft\textbf{Appearance Modeling.}}\quad To handle the appearance variation that often occurs in tracking scenarios, typical discriminative approaches~\cite{stark, mixformer, cui2023mixformerv2} leverage a score model trained to discriminate whether the tracked region contains the target.  Recently, SeqTrack~\cite{SeqTrack} introduced a likelihood-based strategy that uses the likelihood of generated tokens to select dynamic templates without incremental training.
Moreover, the above methods rely on hand-tuned parameters such as update interval and threshold for specific benchmarks~\cite{dai2020high, siamFC, zhu2018distractor}. Furthermore, both of them model appearance in discrete frames~\cite{bhat2019learning, danelljan2019atom, li2019gradnet,wang2020tracking,yang2020roam,yang2018learning,zhang2019learning}. In contrast, we present generative autoregressive appearance reconstruction to model appearance evolution in successive video with end-to-end single-stage training, fully exploiting the temporal potential. 

%% file: 03_method.tex
\section{Method}
\label{sec:method}

\subsection{Revisiting ARTrack}

ARTrack\cite{ARTrack} constitutes a sequence generation framework for visual tracking, with its primary focus centered on the generation of time-series coordinates. This is achieved by utilizing a shared vocabulary to tokenize the object's trajectory, representing it as a discrete sequence of coordinates. Subsequently, the framework employs an encoder-decoder architecture to assimilate visual information and progressively model the sequential evolution of the trajectory prompted by preceding coordinate tokens. This modeling is expressed as a conditional probability:
\begin{equation}
\label{eq:model}
P\left(\boldsymbol{Y}^t|\boldsymbol{Y}^{t-N:t-1},(\bm{C},\boldsymbol{Z},\boldsymbol{X}^t)\right),
\end{equation}
Here, $\boldsymbol{Z}$ and $\boldsymbol{X}^t$ represent the given template and search images at time step $t$, $\bm{C}$ serves as the command token, and $\boldsymbol{Y}$ signifies the target sequence associated with $\boldsymbol{X}$.

Beyond frame-level training and optimization, ARTrack is learned over video sequence with structure objectives to obviate bias between the training and testing phases in data distributions and task objectives. Furthermore, a task-specific SIoU loss\cite{gevorgyan2022siou} is utilized to enhance accuracy.

{\flushleft\textbf{Motivation.}}\quad
At the core of tracking lies the challenge of determining where to focus attention and how to describe the target accurately. ARTrack offers valuable insights by emphasizing the importance of continuous trajectory evolution, allowing for precise ``reading out" of the object's position. However, it falls short in effectively ``retelling" the object's changing appearance over time. Furthermore, ARTrack employs an approach known as intra-frame autoregression, which involves generating four tokens of the bounding box sequentially. This intra-frame autoregression method significantly hampers tracking efficiency. To address these limitations, we introduce ARTrackV2, which leverages joint trajectory-appearance evolution and utilizes a pure encoder architecture to enhance processing speed.

\subsection{Joint Trajectory-Appearance Autoregression}
\begin{figure}
    \centering
    \includegraphics[width=\linewidth]{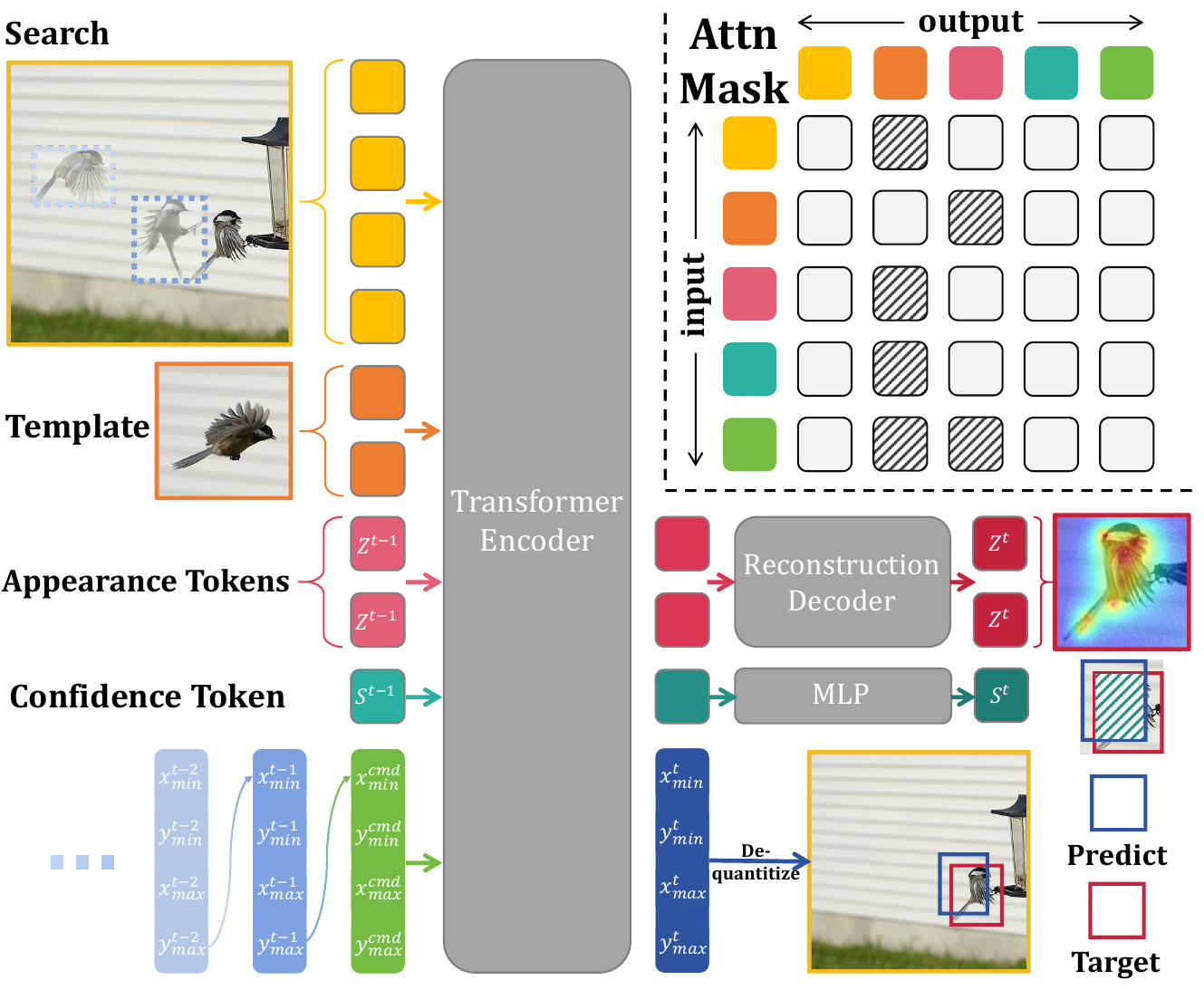}
    \caption{\textbf{ARTrackV2 framework}. Initially, we utilize a Transformer encoder to process all tokens within a frame in parallel, with a masking strategy shown on the top right. Subsequently, appearance tokens are directed to a reconstruction decoder, where the object's appearance within the ongoing search region is reconstructed. Simultaneously, the confidence token is fed into an MLP to predict the IoU between the estimated and ground truth bounding boxes, serving as a measure of the quality of appearance tokens.}
    \label{fig:ARTrackV2}
\end{figure}

The framework of ARTrackV2 is depicted in Figure~\ref{fig:ARTrackV2}.
We expand upon the concept introduced in ARTrack by synchronously modeling the evolution of both trajectory and appearance, thus reinforcing inter-frame autoregression. This is formulated as a probability expression:
\begin{equation}
\label{eq:model}
P\left(\boldsymbol{Y}^t,\textcolor{red}{\boldsymbol{Z}^t}|\boldsymbol{Y}^{t-N:t-1},\textcolor{red}{\boldsymbol{Z}^{t-1}},(\bm{C},\boldsymbol{Z}^{0},\boldsymbol{X}^t)\right),
\end{equation}
Here, $\boldsymbol{Z}^{0}$ represents the initial template, which remains static throughout tracking. The appearance tokens function as dynamic templates, denoted as $\boldsymbol{Z}^{t-1}$ and $\boldsymbol{Z}^{t}$ in \textcolor{red}{red}, effectively describing the temporal evolution of the target's appearance in a continuous manner.

{\flushleft\textbf{Pure Encoder Architecture.}}\quad
In the context of generative paradigm trackers~\cite{ostrack, swin}, there is an efficiency drawback associated with intra-frame autoregression when compared to prevailing tracking methods. Therefore, in our pursuit of simplicity and efficiency, we opt for a transformer encoder~\cite{vit, mae} that can process all tokens within a frame in parallel.
Initially, both the template and search images are divided into patches, flattened, and projected to form a sequence of token embeddings. Similar to ARTrack, we map the object's trajectory across frames to a common coordinate system and tokenize it as trajectory prompts, utilizing a shared vocabulary~\cite{chen2021pix2seq, chen2022unified, wei2023sparse}. We then concatenate visual tokens, trajectory tokens, and four command tokens (each representing one of the four bounding box tokens), add positional and identity embeddings, and input them into the transformer encoder.
This approach eliminates the need for intra-frame autoregression while still maintaining the time-autoregressive nature of the framework, thereby improving overall efficiency.

{\flushleft\textbf{Autoregressive Appearance Reconstruction.}}\quad
We employ a set of appearance tokens alongside a reconstructed decoder to recreate the target's appearance within the current search region. These appearance tokens, termed ``appearance prompts", operate akin to dynamic templates. For each video clip, they initialize as the template within the first frame. In each subsequent frame, they interact with the current search region to extract the target's appearance, through the transformer encoder. Then appearance tokens enter the reconstructed decoder, which rebuilds the target's appearance formed as the feature map of the search region cropped based on the object's position. The output from the reconstructed decoder continuously updates the appearance tokens, which propagates into subsequent frames.
However, a challenge arises when the target becomes invisible, either due to being out of view or significantly occluded. In such instances, invisible appearance propagation can erroneously guide the model to ``read out" nonsensical target localization in the following frames. To address this, we instruct the appearance tokens to maintain their current state in scenarios devoid of visual cues, to prevent unwarranted appearance evolution, ensuring accurate model behavior. This process allows the model to capture appearance variation over time while preserving its autoregressive nature.

{\flushleft\textbf{Appearance Evolution Indicator.}}\quad
In scenarios involving complex conditions such as full occlusion, improper evolution of appearance can result in the loss of target. To tackle this challenge, we propose a solution that guides the model's evolution of appearance with an indicator.
Our approach employs a learnable confidence token and a confidence prediction module comprises a three-layer perceptron. Moreover, we adopt Intersection over Union (IoU) as the indicator's metric based on the fact that it aligns with common tracking evaluation metrics.
In continuous frames, the confidence token interacts with all tokens through the transformer encoder. This interaction implicitly guides the appearance tokens regarding whether to evolve or maintain their current state. Subsequently, the confidence token feeds into the perceptron, predicting the IoU between the model's estimations and the ground truth boxes. Like appearance tokens, the estimated indicator updates the previous confidence token and propagates into the subsequent frame.

{\flushleft\textbf{Oriented Masking.}}\quad
To prevent the model from exclusively fixating on cropping visual features solely based on predicted localization and overlooking the understanding of appearance evolution, we implement an attention masking strategy within the transformer encoder.
Beyond MixFormer~\cite{mixformer}, which concerns intrinsic characteristics within templates to eliminate potential interactive distractors, our approach involves restricting appearance tokens. We force appearance tokens to solely interact with the search region (for reconstructing the target's appearance) and confidence token (for instructing appearance evolution). This deliberate process aims to deter appearance tokens from merely cropping visual features based on target localization and then limits the comprehension of appearance evolution.

\subsection{Sequence Augmentation}
When compared to frame-level training, which involves sampling image pairs from videos~\cite{siamFC, SiamRCNN, li2019siamrpn++}, sequence-level training~\cite{ARTrack, slt} aligns the training and testing data distributions by exclusively sampling successive video clips instead of individual image pairs. However, this approach results in a sharp reduction in the amount of training data available. To overcome this challenge, we investigate sequence-level augmentation methods.

Drawing inspiration from multiple object tracking techniques~\cite{motr, motrv2, motiontrack}, we experiment with fixed and random interval sampling, but both of these methods negatively impact tracking accuracy. We observe that these approaches disrupt the natural progression of temporal information, leading the tracker to learn spurious temporal interactions.

As a result, our criterion for designing augmentation is to preserve the time-series nature of the data. We propose a straightforward yet effective augmentation strategy known as ``reverse augmentation". Given a video sequence, we invert it with a certain probability to expand the training dataset.

\subsection{Training and Inference}
ARTrackV2 emphasizes video sequence-level training and facilitates joint trajectory-appearance evolution in an end-to-end manner.

{\flushleft\textbf{Training.}}\quad
Similar to its predecessor, ARTrackV2 undergoes sequence-level training. We employ a structured objective that maximizes the log-likelihood of trajectory sequences. Moreover, we incorporate a task-agnostic SIoU loss~\cite{gevorgyan2022siou} to enhance the measurement of spatial correlation.

While modeling the trajectory over time, ARTrackV2 also maintains an autoregressive model to reconstruct appearance. Drawing inspiration from MAE~\cite{mae}, we introduce reconstruct tokens masking strategy, after the transformer encoder processing, we sample a subset of appearance tokens and mask them. This creates a challenging task aimed at preventing the overfitting of reconstruction. Subsequently, we compute the mean squared error (MSE) between the reconstructed tokens and the target within the search region or the preceding appearance tokens, depending on whether the object is visible. To avoid poor-quality appearance evolution, we introduce the confidence prediction module, trained by L1 loss between the actual and predicted IoU.

For each video clip, the cached trajectory prompts are initialized with the bounding box from the first frame, and the appearance tokens are set to match the template. Both the trajectory-appearance prompts and the confidence token are iteratively propagated into subsequent frames in an autoregressive manner. The overall tracker is optimized by sequence-level loss function, which is defined as follows:
\begin{equation}
    \label{eq:loss}
\mathcal{L}=\mathcal{L}_\text{ce}+\lambda_\text{SIoU}\mathcal{L}_\text{SIoU}+\lambda_\text{mse}\mathcal{L}_\text{mse}+\lambda_\text{L1}\mathcal{L}_\text{L1},
\end{equation}
where $\mathcal{L}_\text{ce}$, $\mathcal{L}_\text{SIoU}$, $\mathcal{L}_\text{mse}$ and $\mathcal{L}_\text{L1}$ are the cross-entropy loss, SIoU loss, MSE loss, and IoU L1 loss respectively. The $\lambda$ values serve as weights to balance the contribution of each loss term.

{\flushleft\textbf{Inference.}}\quad During inference, we initialize the trajectory and appearance tokens as previously described. Subsequently, we simultaneously generate the trajectory sequence from the estimated likelihood with $\texttt{argmax}$ sampling and reconstruct the target's appearance. This process is carried out in an autoregressive manner, where the trajectory, appearance, and confidence token are iteratively propagated into subsequent frames. It's worth noting that, during the reconstruction process, we intentionally refrain from appearance tokens masking, unlike the training phase.

%% file: 04_experiments.tex
\section{Experiments}

\label{sec:experiments}

\subsection{Implementation Details}
The models train with 8 NVIDIA RTX A6000 GPUs, with training times ranging from approximately 26 to 120 hours, depending on the specific experimental configurations.

{\flushleft\textbf{Model Variants.}}\quad
We trained three variants of ARTrackV2 with different configurations as follows:
\begin{itemize}
\item \textbf{$\text{ARTrackV2}_\text{256}$.} Backbone: ViT-Base; Template size: [128\begin{math}\times\end{math}128]; Search region size: [256\begin{math}\times\end{math}256];
\item \textbf{$\text{ARTrackV2}_\text{384}$.} Backbone: ViT-Base; Template size: [192\begin{math}\times\end{math}192]; Search region size: [384\begin{math}\times\end{math}384];
\item \textbf{$\text{ARTrackV2-L}_\text{384}$.} Backbone: ViT-Large; Template size: [192\begin{math}\times\end{math}192]; Search region size: [384\begin{math}\times\end{math}384].
\end{itemize}

{\flushleft\textbf{Training Strategy.}}\quad
We adhere to established protocols for training and evaluating our models, consistent with ARTrack. The training dataset includes GOT-10k~\cite{huang2019got} (with 1k sequences removed from the GOT-10k train split, as per~\cite{stark}), TrackingNet~\cite{muller2018trackingnet}, and LaSOT~\cite{fan2019lasot}.
To ensure a fair evaluation of the GOT-10k test set, our models learn from the entire GOT-10k training split following its one-shot protocol.

The models are optimized using AdamW~\cite{weight_decay} with a weight decay of $5\times10^{-2}$. The learning rate for the backbone is set to $8\times10^{-6}$, while other parameters use a learning rate of $8\times10^{-5}$.
The training process comprises 60 epochs, with 960 video sequences in each epoch. Each sequence consists of 32 frames, constrained by GPU memory limitations.

Furthermore, in line with ARTrack~\cite{ARTrack}, and to align with established trackers~\cite{tatrack, cttrack, SeqTrack, GRM} that are trained using image datasets such as COCO2017~\cite{lin2014microsoft}, we employ frame-level training to pre-train our models. During this process, we utilize four training datasets and apply image data augmentation techniques, including horizontal flip and brightness jittering, which are consistent with OSTrack~\cite{ostrack} and SeqTrack~\cite{SeqTrack}. The pre-trained models are optimized using AdamW with a weight decay of $10^{-4}$, with the learning rate for the backbone and other parameters set the same as previously mentioned. Our pre-trained model undergoes 240 epochs of training, with 60k matching pairs processed per epoch.

\subsection{Main Results}

\begin{table*}[th]
  \centering
  \setlength{\tabcolsep}{4pt}
  \resizebox{\linewidth}{!}{
  \begin{NiceTabular}{r|ccc|ccc|ccc|ccc}
    \toprule
    \multirow{2}{*}{Methods} &
    \multicolumn{3}{c|}{GOT-10k*} & \multicolumn{3}{c|}{TrackingNet} &\multicolumn{3}{c|}{LaSOT} &  \multicolumn{3}{c}{LaSOT\footnotesize{ext}}  \\
    \cmidrule{2-13}
     & AO(\%) & $SR_{0.5}$(\%) & $SR_{0.75}$(\%)& AUC(\%) & $P_{Norm}$(\%) & $P$(\%) & AUC(\%) & $P_{Norm}$(\%) & $P$(\%) & AUC(\%) & $P_{Norm}$(\%) & $P$(\%) \\
    \midrule
    $\text{SiamFC}_\text{255}$~\cite{siamFC} & 34.8 & 35.3 & 9.8 & 57.1 & 66.3 & 53.3  & 33.6 & 42.0 & 33.9 & 23.0 & 31.1 & 26.9 \\
    $\text{ECO}_\text{224}$~\cite{eco}& 31.6 & 30.9 & 11.1 & 55.4 & 61.8 & 49.2 & 32.4 & 33.8 & 30.1 & 22.0 & 25.2 & 24.0  \\
    $\text{DiMP}_\text{288}$~\cite{DiMP}& 61.1 & 71.7 & 49.2 & 74.0 & 80.1 & 68.7 & 56.9 & 65.0 & 56.7 & 39.2 & 47.6 & 45.1\\
    $\text{SiamR-CNN}_\text{255}$~\cite{SiamRCNN}& 64.9 & 72.8 & 59.7 & 81.2 & 85.4 & 80.0 & 64.8 & 72.2 & - & - & - & - \\
    $\text{Ocean}_\text{255}$~\cite{Ocean}& 61.1 & 72.1 & 47.3& - & - & -  & 56.0 & 65.1 & 56.6 & - & - & - \\
    $\text{TrDiMP}_\text{352}$~\cite{TrDiMP}& 67.1 & 77.7 & 58.3& 78.4 & 83.3 & 73.1  & 63.9 & - & 61.4 & - & - & - \\
    $\text{SLT-TrDiMP}_\text{352}$~\cite{slt} & 67.5 & 78.8 & 58.7 & 78.1 & 83.1 & 73.1 & 64.4 & 73.5 & - & - & - & - \\
    $\text{TransT}_\text{256}$~\cite{TransT}& 67.1 & 76.8 & 60.9 & 81.4 & 86.7 & 80.3 & 64.9 & 73.8 & 69.0 & - & - & -\\
    $\text{STARK}_\text{320}$~\cite{stark}& 68.8 & 78.1 & 64.1 & 82.0 & 86.9 & -  & 67.1 & 77.0 & - & - & - & -\\
    $\text{SwinTrack-B}_\text{384}$~\cite{swin}& 72.4 & 80.5 & 67.8 & 84.0 & - & 82.8 & 71.3 & - & 76.5 & 49.1 & - & 55.6 \\
    $\text{MixFormer-L}_\text{320}$~\cite{mixformer}& - & - & - & 83.9 & {88.9} & 83.1 & 70.1 & 79.9 & 76.3 & - & - & -\\
    $\text{OSTrack}_\text{384}$~\cite{ostrack}& {73.7} & {83.2} & {70.8}  & 83.9 & 88.5 & 83.2 & {71.1} & {81.1} & {77.6} & {50.5} & {61.3} & {57.6}\\
    $\text{CTTrack-B}_\text{320}$~\cite{cttrack} & 71.3 & 80.7 & 70.3 & 82.5 & 87.1 & 80.3 & 67.8 & 77.8 & 74.0 & - & - & - \\
    $\text{CTTrack-L}_\text{320}$~\cite{cttrack} & 72.8 & 81.3 & 71.5 & 84.9 & 89.1 & 83.5 & 69.8 & 79.7 & 76.2 & - & - & - \\
    $\text{TATrack-B}_\text{224}$~\cite{tatrack} & 73.0 & 83.3 & 68.5 & 83.5 & 88.3 & 81.8 & 69.4 & 78.2 & 74.1 & - & - & - \\
    $\text{TATrack-L}_\text{384}$~\cite{tatrack} & - & - & - & 85.0 & 89.3 & 84.5 & 71.1 & 79.1 & 76.1 & - & - & - \\
    $\text{GRM-B}_\text{256}$~\cite{GRM} & 73.4 & 82.9 & 70.4 & 84.0 & 88.7 & 83.3 & 69.9 & 79.3 & 75.8 & - & - & - \\
    $\text{GRM-L}_\text{320}$~\cite{GRM} & - & - & - & 84.4 & 88.9 & 84.0 & 71.4 & 81.2 & 77.9 & - & - & - \\
    $\text{MixViT}_\text{288}$~\cite{mixformer} & 72.5 & 82.4 & 69.9 & 83.5 & 88.3 & 82.0 & 69.6 & 79.9 & 75.9 & - & - & - \\
    $\text{MixViT-L}_\text{384}$~\cite{mixformer} & 75.7 & 85.3 & 75.1 & 85.4 & \underline{90.2} & 85.7 & 72.4 & \underline{82.2} & \uwave{80.1} & - & - & - \\
    $\text{SeqTrack-B}_\text{256}$~\cite{SeqTrack} & 74.7 & 84.7 & 71.8 & 83.3 & 88.3 & 82.2 & 69.9 & 79.7 & 76.3 & 49.5 & 60.8 & 56.3 \\
    $\text{SeqTrack-L}_\text{384}$~\cite{SeqTrack} & 74.8 & 81.9 & 72.2& 85.5 & \uwave{89.8} & \uwave{85.8} & 72.5 & 81.5 & 79.3 & 50.7 & 61.6 & 57.5 \\
    \midrule
    $\text{ARTrack}_\text{256}$~\cite{ARTrack} & 73.5 & 82.2 & 70.9 & {84.2} & 88.7 & {83.5} & 70.4 & 79.5 & 76.6 & 46.4 & 56.5 & 52.3 \\
    $\text{ARTrack}_\text{384}$~\cite{ARTrack} & 75.5 & 84.3 & 74.3 & 85.1 & 89.1 & 84.8 & 72.6 & 81.7 & 79.1 & 51.9 & 62.0 & 58.5 \\
    $\text{ARTrack-L}_\text{384}$~\cite{ARTrack} & \underline{78.5} & \underline{87.4} & \underline{77.8}& \uwave{85.6} & 89.6& \underline{86.0} & \underline{73.1} & \underline{82.2} & \underline{80.3} & \uwave{52.8} & \uwave{62.9} & \underline{59.7} \\
    \textbf{$\text{ARTrackV2}_\text{256}$} & 75.9 & 85.4 & 72.7 & 84.9 & 89.3 & 84.5 & 71.6 & 80.2 & 77.2 & 50.8 & 61.9 & 57.7 \\

    \textbf{$\text{ARTrackV2}_\text{384}$} & \uwave{77.5} & \uwave{86.0} & \uwave{75.5} & \underline{85.7} & \uwave{89.8} & 85.5 & \uwave{73.0} & \uwave{82.0} & 79.6 & \underline{52.9} & \underline{63.4} & \uwave{59.1} \\

    \textbf{$\text{ARTrackV2-L}_\text{384}$} & \bf{79.5} & \bf{87.8} & \bf{79.6} & \bf{86.1} & \bf{90.4} & \bf{86.2} & \bf{73.6} & \bf{82.8} & \bf{81.1} & \bf{53.4} & \bf{63.7} & \bf{60.2} \\
    \bottomrule
  \end{NiceTabular}
  }
  \caption{State-of-the-art comparison on GOT-10k~\cite{huang2019got}, TrackingNet~\cite{muller2018trackingnet}, LaSOT~\cite{fan2019lasot} and {LaSOT\footnotesize{ext}}~\cite{fan2021lasot}. Where * denotes only trained on GOT-10k. The number in the subscript denotes the search region resolution. Best in \textbf{bold}, second best \underline{underlined}, and third best \uwave{underwave}.}
  \label{tab:main}
\end{table*}

We evaluate the performance of our proposed $\text{ARTrackV2}_\text{256}$ and $\text{ARTrackV2-L}_\text{384}$ on several benchmarks, including
GOT-10k~\cite{huang2019got}, TrackingNet~\cite{muller2018trackingnet}, LaSOT~\cite{fan2019lasot} and {LaSOT\footnotesize{ext}}~\cite{fan2021lasot}. 

{\flushleft\textbf{GOT-10k~\cite{huang2019got}.}}\quad GOT-10k is a comprehensive generic object tracking dataset comprising video sequences featuring real-world moving objects with manually annotated bounding boxes. The dataset advocates for a one-shot protocol, which necessitates that trackers are trained exclusively on the GOT-10k training split to ensure that the object classes in the training and testing sets do not overlap. Adhering to this protocol, our ARTrackV2 is trained exclusively on the GOT-10k training split and evaluated on the test set. As demonstrated in Table \ref{tab:main}, our $\text{ARTrackV2-L}_\text{384}$ outperforms state-of-the-art trackers across all metrics. Notably, our $\text{ARTrackV2}_\text{256}$ and  $\text{ARTrackV2}_\text{384}$ surpasses other trackers with higher resolution and larger backbones except ARTrack.

{\flushleft\textbf{TrackingNet~\cite{muller2018trackingnet}.}}\quad TrackingNet is an extensive tracking dataset comprising over 30,000 videos that cover a wide range of real-world scenarios and content. Each video is annotated with manually labeled bounding boxes. We assess the performance of ARTrackV2 on its test set which contains 511 videos covering diverse object categories, as illustrated in Table \ref{tab:main}. This table shows that not only does our $\text{ARTrackV2}_\text{384}$ outperform all other trackers in AUC, but our $\text{ARTrackV2-L}_\text{384}$ also establishes a new state-of-the-art in three matrices on this large-scale benchmark.

{\flushleft\textbf{LaSOT~\cite{fan2019lasot}.}}\quad LaSOT is a benchmark designed for long-term tracking, comprising 280 videos in its test set, effectively assessing the tracker's robustness in extended video sequences. Table \ref{tab:main} demonstrates that our $\text{ARTrackV2}_\text{256}$ achieves comparable performance to $\text{ARTrack}_\text{384}$, despite having lower input resolution. Furthermore, our $\text{ARTrackV2-L}_\text{384}$ significantly enhances performance, setting a new state-of-the-art while running at 49 FPS, which is over $5\times$ faster than $\text{SeqTrack-L}_\text{384}$ (9 FPS).

{\flushleft\textbf{LaSOT{\footnotesize{ext}}~\cite{fan2021lasot}.}}\quad
{LaSOT\footnotesize{ext}} serves as an extension of LaSOT, including an additional 150 videos. These new sequences introduce challenging tracking scenarios, such as occlusions and variations in small objects. To demonstrate the robustness of our models in handling these difficult scenarios, we evaluate ARTrackV2 and present the results in Table \ref{tab:main}. Remarkably, our ARTrackV2\textsubscript{384} achieves exceptional accuracy, surpassing other trackers with larger backbones with 72 FPS. Furthermore, our ARTrackV2-L\textsubscript{384} outperforms ARTrack-L\textsubscript{384} and establishes a new state-of-the-art performance, operating at a speed of 49 FPS, which is $3\times$ faster than ARTrack.

\begin{figure}
    \centering
    \includegraphics[width=\linewidth]{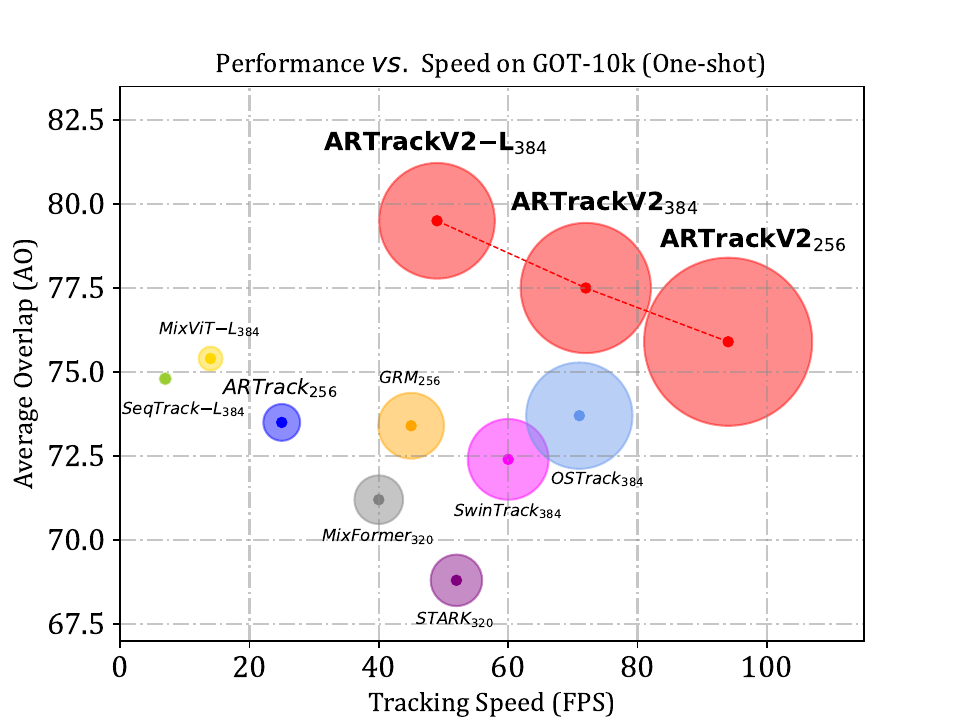}
    \caption{Comparison of accuracy vs. latency trade-off for different tracking methods in GOT-10k (one-shot setting). }
    \label{fig:tradeoff}
    \vspace{-4pt}
\end{figure}

\subsection{Accuracy vs. Latency}
In comparison to ARTrack, we have eliminated intra-frame autoregression to enhance inference speed, resulting in nearly a $3\times$ improvement in inference efficiency without compromising accuracy. To illustrate this improvement, we conducted a comparative analysis of state-of-the-art trackers on GOT-10k using the one-shot protocol, as depicted in Figure \ref{fig:tradeoff}. Our ARTrackV2-L\textsubscript{384} achieves a new state-of-the-art with an impressive AO of 79.5\%, while ARTrackV2\textsubscript{256} delivers competitive performance at 94 FPS, surpassing other trackers with higher resolutions and larger backbones.

\subsection{Experimental Analyses}
We analyze the main properties of the ARTrackV2. For the following experimental studies, we follow the GOT-10k test protocol unless otherwise noted. Default settings are marked in \colorbox{gray!25}{gray}.

\def\arraystretch{1.0}
\renewcommand{\tabcolsep}{4 pt}
\begin{table}[h]\footnotesize
  \centering
  \begin{NiceTabular}{l|ccc|c}
    \toprule

    model variants & AO & $SR_{0.5}$ & $SR_{0.75}$ & FPS  \\
    \midrule
    ARTrack~\cite{ARTrack} & 73.5 & 82.2 & 70.9 & 26 \\
    \midrule
    pure encoder architecture & 71.0 & 79.9 & 68.2 & 116 \\
    $+$ appearance evolution & 74.2 & 83.1 & 71.4 & 98 \\
    $+$ confidence prediction & 74.7 & 83.7 & 72.1 & 94 \\
    $+$ masking strategy & 75.2 & 84.8 & 72.4 & 94\\   
    $+$ sequence augmentation & \default{\textbf{75.9}} & \default{\textbf{85.4}}  & \default{\textbf{72.7}}& \default{\textbf{94}} \\
    \bottomrule
  \end{NiceTabular}
  \caption{Summary of \textbf{cumulative effects}.}
  \label{tab:summary}
  \vspace{-4pt}
\end{table}

{\flushleft\textbf{Summary of Cumulative Effect.}}\quad
We conduct comprehensive ablation studies to analyze our proposed approach, considering several key aspects: evaluating the impact of the pure encoder architecture, assessing the effectiveness of autoregressive appearance evolution, examining the contribution of the confidence prediction module, validating the masking strategy, and exploring the benefits of sequence-level data augmentation. Additionally, we systematically evaluate the cumulative effects of integrating these various components, and the results are summarized in Table \ref{tab:summary}.

We observed that adopting the pure encoder architecture significantly improved tracking efficiency. However, this improvement came at the cost of a decrease in accuracy, which we attributed to the lack of intra-frame temporal information. To address this, we introduced autoregressive appearance evolution which leveraged appearance reconstructions, confidence prediction, attention masking strategy, and sequence data augmentation to strengthen inter-frame autoregression. These modifications teach the model to ``retell" the object's appearance variation in a time-autoregressive manner. As a result of these enhancements, we complement generative paradigm trackers to joint evolution of trajectory and appearance, achieving substantial improvements in model performance, and ultimately establishing state-of-the-art results.

\def\arraystretch{1.0}
\renewcommand{\tabcolsep}{3 pt}
\begin{table}[h]\footnotesize
  \centering
\begin{NiceTabular}{c|cc|c}
    \toprule
    appearance model & single-stage & thresholds tuning & AO\\
    \midrule
    discriminative (score-based) & \ding{55} & \ding{51} & 74.5\\
    discriminative (likelihood-based) & \ding{51} & \ding{51} & 74.1\\
    \textbf{generative (reconstruction)} & \ding{51} & \ding{55} & \default{\textbf{75.9}}\\
    \bottomrule
\end{NiceTabular}
  \caption{\textbf{Appearance model comparison}.}
  \label{tab:update}
  \vspace{-4pt}
\end{table}

\begin{figure}[!t]
    \centering
    \includegraphics[width=1.0\linewidth]{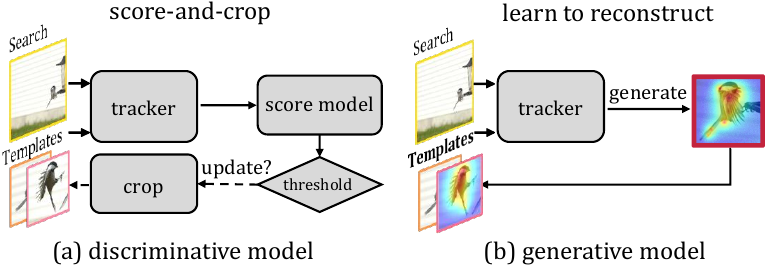}
    \caption{\textbf{Comparison of appearance modeling approaches}. (a) discriminative model adopts a score-and-crop strategy to decide updates. (b) generative model learns to reconstruct the template.}
    \label{fig:update_compare}
    \vspace{-5pt}
\end{figure}

{\flushleft\textbf{Appearance Model.}}\quad
In this subsection, we delve into the generative appearance model in ARTrackV2. This model differs from previous discriminative models, which determine whether the cropped region from the search image, using the tracking result, is reliable for updating the template. Discriminative approaches typically require an additional stage to train a score model~\cite{stark, mixformer, cui2023mixformerv2} to classify whether the tracked region contains the target object, thus breaking the single-stage end-to-end learning framework. SeqTrack introduces a likelihood-based strategy that uses the likelihood of the generated coordinate tokens to select dynamic templates in a single stage. Moreover, these methods often involve hand-tuning parameters for each individual dataset, including score/likelihood thresholds and update frequency.
As depicted in Figure \ref{fig:update_compare}, ARTrackV2 employs a simple unified approach to appearance evolution. Instead of score-and-crop, it learns to recreate the template in a continuous autoregressive manner. It also adopts a masking strategy to prevent attention from appearance tokens to the trajectory ones. We compare these different appearance models in Table \ref{tab:update}, demonstrating that our generative model performs better than score-based or likelihood-based discriminative approaches.

\def\arraystretch{1.0}
\renewcommand{\tabcolsep}{4 pt}
\begin{table}[h]\footnotesize
  \centering
\begin{NiceTabular}{c|ccc}
    \toprule
    reconstruction objective & AO & $SR_{0.5}$ & $SR_{0.75}$ \\
    \midrule
    image reconstruction & 74.8 & 83.9 & 71.1 \\
    \textbf{feature reconstruction} & \default{\textbf{75.9}} & \default{\textbf{85.4}} & \default{\textbf{72.7}} \\
    \bottomrule
\end{NiceTabular}
  \caption{\textbf{Reconstruction objective}.}
  \label{tab:reconstruct}
  \vspace{-4pt}
\end{table}

{\flushleft\textbf{Reconstruction Objective.}}\quad
The quality of the appearance tokens is ascertained through an adequate reconstruction objective of the target, which can be in the image pixel domain or in the latent feature domain. To investigate this, we conducted exploratory experiments, as outlined in Table \ref{tab:reconstruct}. Our findings indicate that ``feature reconstruction" leads to a noteworthy improvement of approximately 1.1\% on the AO metric. This improvement demonstrates that feature reconstruction is more effective in appearance evolution. In contrast, image reconstruction may tend to excessively focus on intricate details or background information. In scenarios characterized by motion blur and occlusion settings, this approach encounters challenges in accurately reconstructing the target at the pixel level.

\def\arraystretch{1.0}
\renewcommand{\tabcolsep}{4 pt}
\begin{table}[h]\footnotesize
  \centering
\begin{NiceTabular}{c|ccc}
    \toprule
    indicator metric  & AO & $SR_{0.5}$ & $SR_{0.75}$  \\
    \midrule
    w/o & 75.1 & 84.7 & 71.9\\
    confidence & 74.9 & 84.4 & 71.5 \\
    distance & 75.1 & 84.5 & 72.2\\
    visibility & 74.6 & 84.2 & 71.3\\
    IoU & \default{\textbf{75.9}} & \default{\textbf{85.4}} & \default{\textbf{72.7}}\\
    \bottomrule
\end{NiceTabular}
  \caption{\textbf{Appearance evolution indicator}.}
  \label{tab:indicator}
  \vspace{-10pt}
\end{table}

{\flushleft\textbf{Appearance Evolution Indicator.}}\quad 
To ensure quality appearance evolution, it is necessary to employ indicators that guide the reconstruction of appearance tokens. These indicators serve to characterize the evolution quality. In our analysis of different metrics' impact on the model, we present the findings in Table \ref{tab:indicator}. The metric labeled ``confidence" ~\cite{stark, mixformer} denotes the confidence assigned to the current template, indicating whether it contains the target. The ``distance" \cite{fu2021stmtrack} represents the cosine distance between the features of the appearance tokens and the target's appearance feature within the ongoing search region. The ``visibility"~\cite{huang2019got} quantifies the visibility ratio of the target in the search region. Lastly, the ``IoU" metric measures the Intersection over Union between the predicted and the ground truth bounding boxes.

Contrary to previous research findings~\cite{dai2020high, fu2021stmtrack, Zhang_2019_ICCV, Guo_2017_ICCV, stark, mixformer}, our investigation has revealed that employing IoU as the reconstruction indicator leads to superior accuracy. This observation can be primarily attributed to the fact that when compared to alternative methods, the IoU metric is more closely aligned with the evaluation metric utilized for tracking. Consequently, it provides a precise reflection of the quality of evolution.
We also note that the visibility metric yields unsatisfactory results. GOT-10k provides an assessment of object visibility segmented into 9 levels, but the boundaries between these levels are often vague and may contain noisy labels. This poses a challenge for models in precisely evaluating the target's visibility.

\def\fwidth{0.18\linewidth}
\def\arraystretch{0.5}
\renewcommand{\tabcolsep}{0.5 pt}
\begin{figure}[t]
\centering
\begin{tabular}{cccccc}
     \rotatebox{90}{\scriptsize{Variation}} \hspace{1pt} &
     \includegraphics[width=\fwidth]{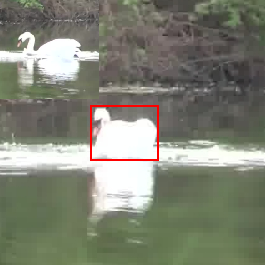} &  
     \includegraphics[width=\fwidth]{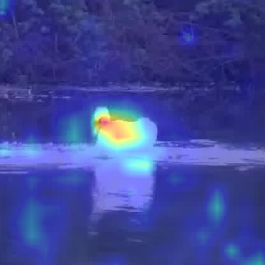} &
     \includegraphics[width=\fwidth]{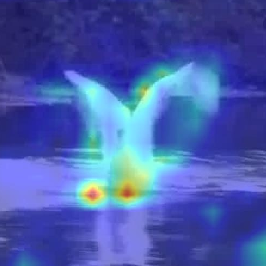} & 
     \includegraphics[width=\fwidth]{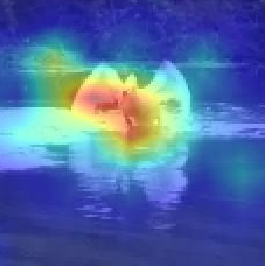} &
     \includegraphics[width=\fwidth]{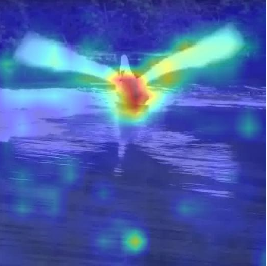} \\
     \rotatebox{90}{\scriptsize{Occlusion}} \hspace{1pt} &
     \includegraphics[width=\fwidth]{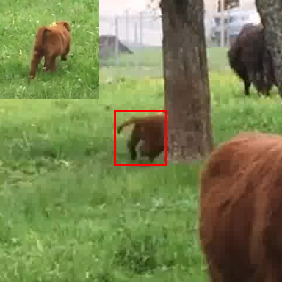} &  
     \includegraphics[width=\fwidth]{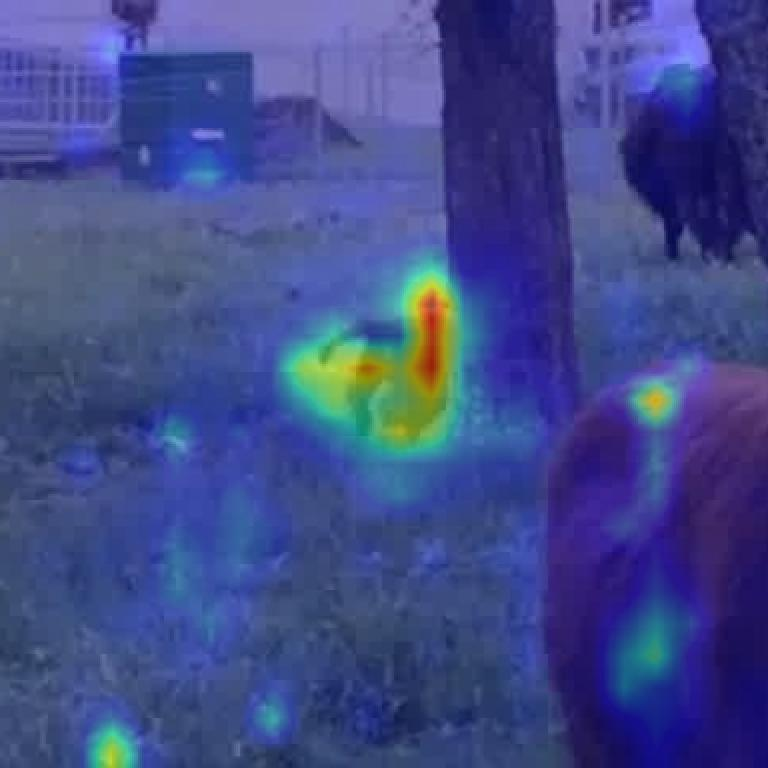} &
     \includegraphics[width=\fwidth]{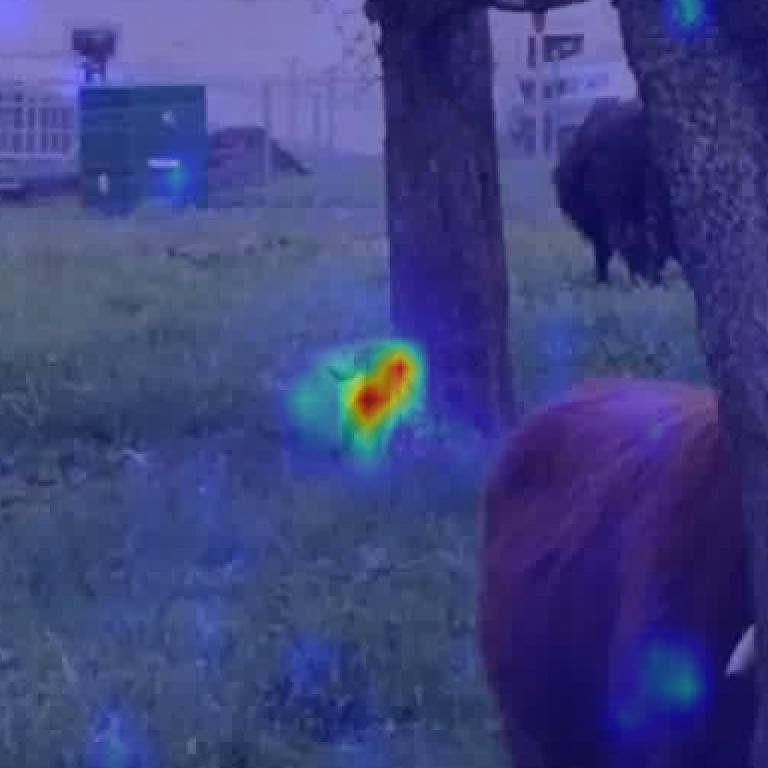} & 
     \includegraphics[width=\fwidth]{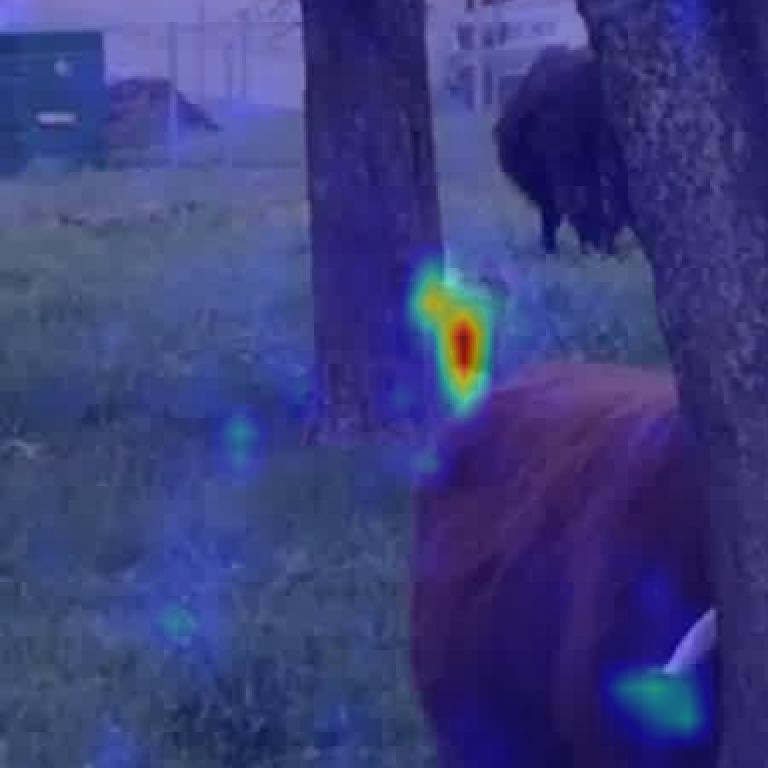} &
     \includegraphics[width=\fwidth]{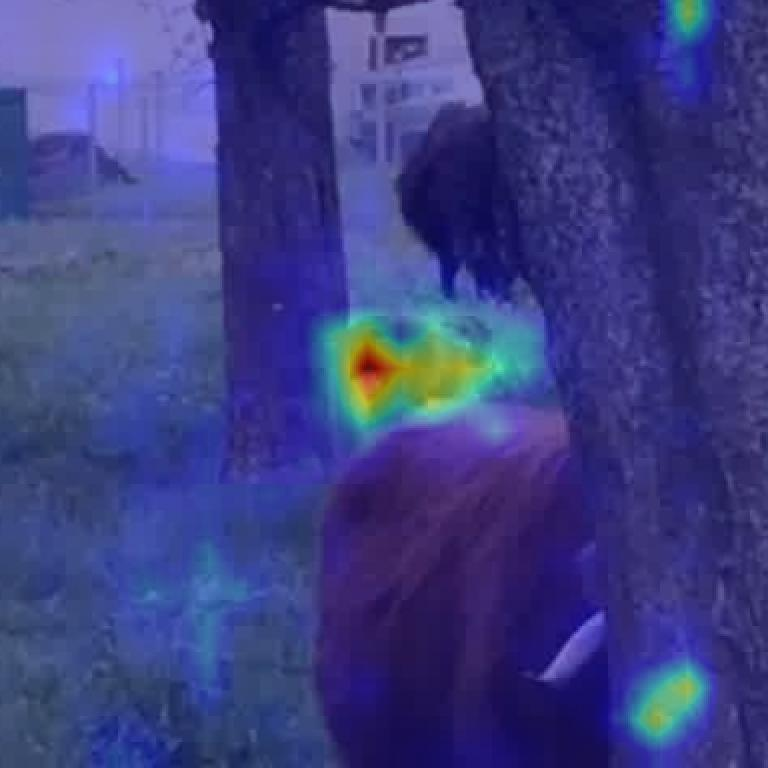} \\
     \rotatebox{90}{\scriptsize{Distraction}} \hspace{1pt}&
     \includegraphics[width=\fwidth]{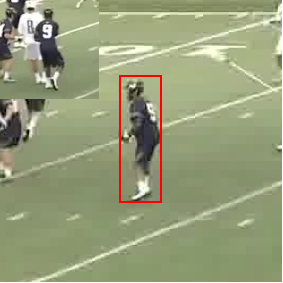} &  
     \includegraphics[width=\fwidth]{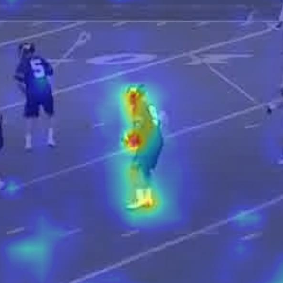} &
     \includegraphics[width=\fwidth]{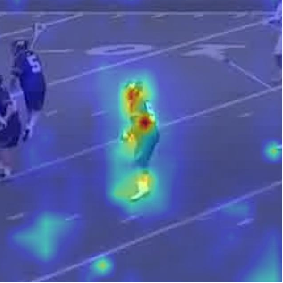} & 
     \includegraphics[width=\fwidth]{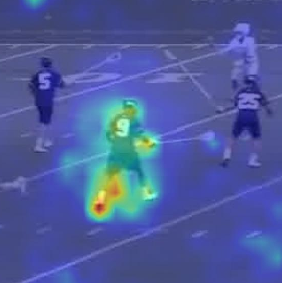} &
     \includegraphics[width=\fwidth]{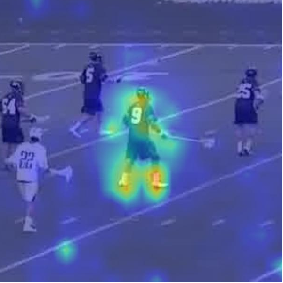} \\
     \rotatebox{90}{\scriptsize{Motion blur}} \hspace{1pt} &
     \includegraphics[width=\fwidth]{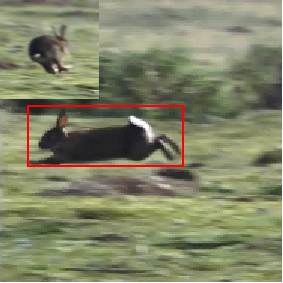} &  
     \includegraphics[width=\fwidth]{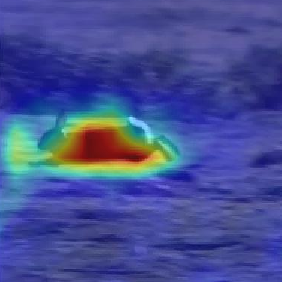} &
     \includegraphics[width=\fwidth]{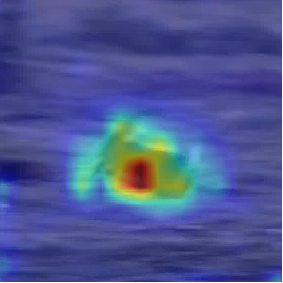} & 
     \includegraphics[width=\fwidth]{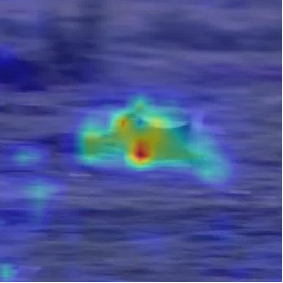} &
     \includegraphics[width=\fwidth]{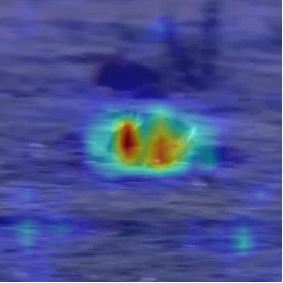} \\
     & \scriptsize{(a) search} & \scriptsize{(b) time $t$} & \scriptsize{(c) time $t+1$} & \scriptsize{(d) time $t+2$} & \scriptsize{(e) time $t+3$} \\ 
\end{tabular}
\caption{\textbf{Attention visualization}. (a): Search region and template. The \textcolor{red}{red} boxes denote the ground truth. (b)-(e): Appearance tokens to search the cross-attention map of ARTrackV2.}
\label{fig:challenge}
\end{figure}

{\flushleft\textbf{Visualization and Analysis.}}\quad
To gain deeper insights into the autoregressive appearance evolution, we generate cross-attention maps about the appearance tokens to the search region while evolving trajectory and appearance. In order to demonstrate the versatility of our model, we challenge it with complex scenarios that pose significant tracking difficulties, including appearance variation, partial occlusion, distribution, and motion blur, as shown in Figure \ref{fig:challenge}. Perceptibly, our model adeptly captures the appearance evolution in successive frames within each of these challenging scenarios. 

When confronted with rapid changes in target appearance and partial occlusions, traditional trackers tend to respond inadequately to these short-term variations. Furthermore, incorrect updates to the target's appearance in such scenarios can render the tracker agnostic to the target in subsequent frames. Our approach leverages the mutual complementation of both appearance and trajectory evolution, consecutively pinpointing the object's location. Joint evolution constructs a more comprehensive representation of the target, thus strengthening inter-frame autoregression in scenarios with incoherent visual or motion cues.

\def\arraystretch{1.0}
\renewcommand{\tabcolsep}{4 pt}
\begin{table}[h]\footnotesize
  \centering
\begin{NiceTabular}{c|ccc}
    \toprule
    sequence augmentation & AO & $SR_{0.5}$ & $SR_{0.75}$ \\
    \midrule
    fixed interval & 74.8 & 84.6 & 71.3 \\
    random interval & 75.1 & 84.9 & 70.9 \\
    \textbf{reverse video} & \default{\textbf{75.9}} & \default{\textbf{85.4}} & \default{\textbf{72.7}} \\
    \bottomrule
\end{NiceTabular} 
  \caption{\textbf{Sequence augmentation comparison}.}
  \label{tab:aug}
  \vspace{-8pt}
\end{table}

{\flushleft\textbf{Sequence Augmentation.}}\quad Sequence-level data augmentation\cite{motionaware_augment, videomix} is a widely used technique in video tasks, but it is rarely employed in visual tracking due to the prevalent use of frame-level training. In contrast, we embrace sequence-level training, where models are trained using video clips instead of image pairs. Therefore, it is necessary to explore video data augmentation, as demonstrated in Table \ref{tab:aug}. We experimented with sampling the video at fixed or random intervals to augment the training data. Unfortunately, both approaches led to a decrease in precision as they disrupted temporal continuity. In contrast, reverse augmentation, which simply plays the video backward, maintains the data distribution well. This straightforward method increases the AO in GOT-10k by 0.7\%.

%% file: 10_conclusion.tex
\section{Conclusion}
\label{sec:discussion}
We introduce ARTrackV2, an end-to-end tracker that extends the concepts of the predecessor by implementing a unified generative framework that jointly evolves trajectory and reconstructs appearance. In continuous time-series, ARTrackV2 simultaneously ``reads out" the target's localization and ``retells"  appearance variation. Then propagating and forecasting trajectory-appearance prompts into successive frames strengthens inter-frame autoregression. Moreover, we employ a pure encoder architecture that enables parallel processing of all tokens within a frame and eliminates less efficient intra-frame autoregression. ARTrackV2 demonstrates remarkable advancements in both performance and efficiency. In the future, we consider extending this unified framework to encompass multiple video-related tasks.

%% file: 12_appendix.tex
\begin{center}
    \centering
    \begin{tabular}{c}
         \includegraphics[width=0.46\textwidth]{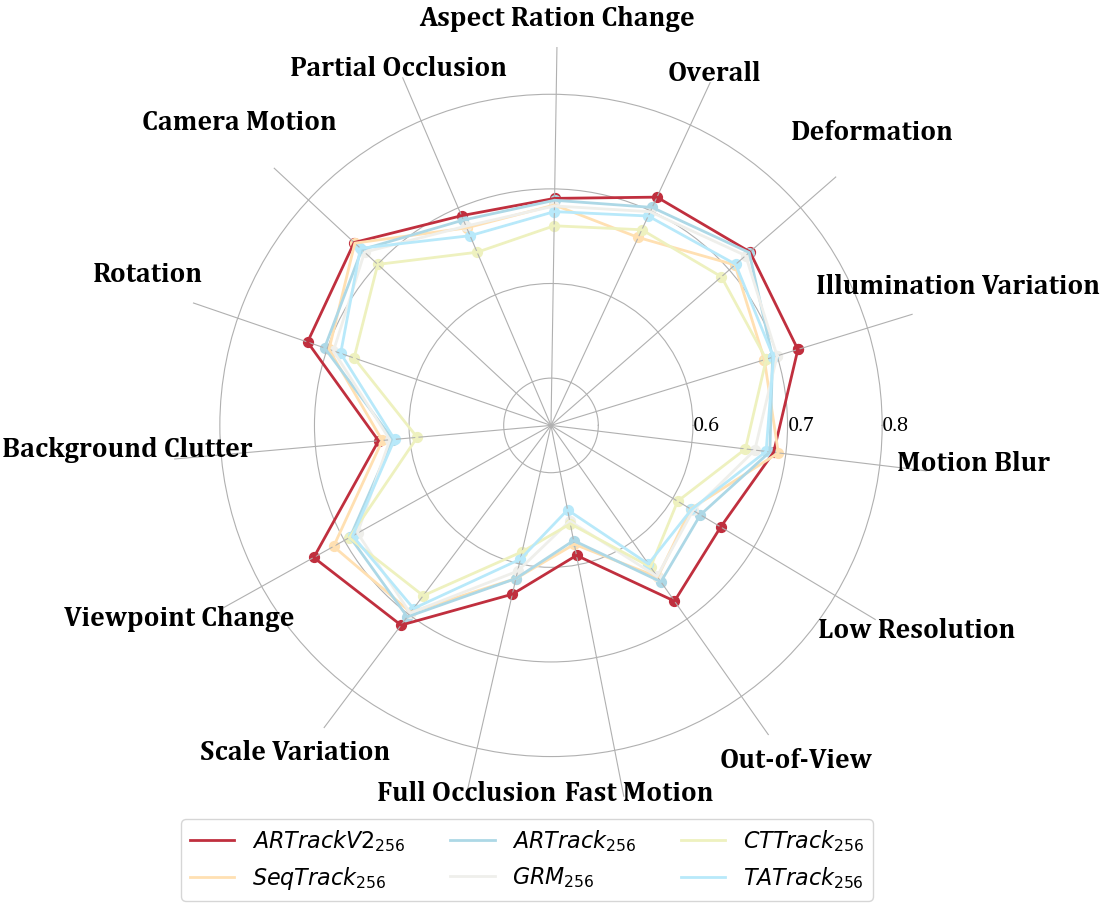}\label{fig:attribute_sota_B} \\ 
         (a) Comparison of base models\vspace{3mm} \\
         \includegraphics[width=0.46\textwidth]{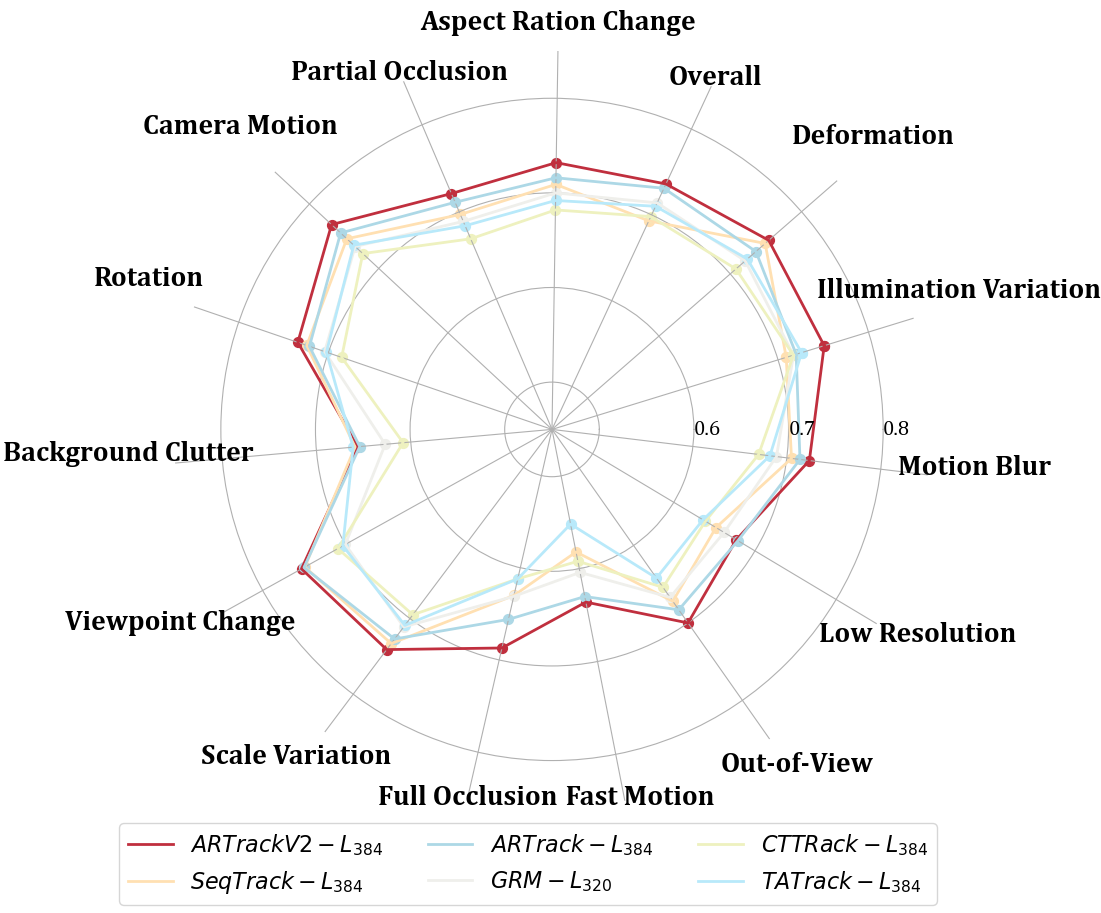}\label{fig:attribute_sota_L} \\
          (b) Comparison of large models
    \end{tabular}   
    \captionof{figure}{Comparison to other state of the art trackers, using the AUC score under different attributes of LaSOT~\cite{fan2019lasot} test set.}
    \label{fig:attribute}
\end{center}

\appendix
\label{sec:appendix}

\section{LaSOT Attributes Results}
LaSOT~\cite{fan2019lasot} benchmark additionally offers a comprehensive range of specific attributes for each video sequence, such as Out-of-View, Full Occlusion, Viewpoint Change, and so on. 
We evaluate the AUC score under various attributes separately, presenting a comparative analysis between ARTrackV2 and other state-of-the-art trackers as shown in Figure~\ref{fig:attribute}(a) and Figure~\ref{fig:attribute}(b).
ARTrackV2 demonstrates noteworthy advancements, particularly in handling scenarios involving Full Occlusion, Partial Occlusion, and Illumination Variation. These attributes necessitate continuous appearance evolution for robust and reliable tracking.

\section{Performance on TNL2K~\cite{wang2021tnl}, NFS~\cite{kiani2017need} and UAV123~\cite{mueller2016uav}}
\input{figs/tnl2k_nfs_uav123}
To showcase the robustness of our ARTrackV2 in a broader spectrum of scenarios, we conducted evaluations on three additional benchmarks: 1) TNL2K, is a multimodal dataset that incorporates natural language tagging and comprises videos demonstrating pedestrian scenarios with cloth/face alterations; 2) NFS, is a dataset that relies on high-speed photography, offering videos at a higher frame rate (240 FPS); 3) UAV123, is a aerial dataset contains long-term videos in the complex scene. Our ARTrackV2-L\textsubscript{384} consistently outperforms other trackers across these benchmarks, as illustrated in Figure \ref{fig:another_three}. Notably, our models achieve this heightened performance without compromising on inference speed.

\section{Sequence Format Study}
\def\arraystretch{1.0}
\renewcommand{\tabcolsep}{4 pt}
\begin{table}[h]\footnotesize
  \centering
  \begin{NiceTabular}{l|ccc}
    \toprule
    sequence form & AO(\%) & $SR_{0.5}$(\%) & $SR_{0.75}$(\%) \\
    \midrule
    \text{[\bm{$x_\text{min}$}$,$ \bm{$y_\text{min}$}$,$ \bm{$x_\text{max}$}$,$ \bm{$y_\text{max}$}]} & \default{\textbf{75.9}} & \default{\textbf{85.4}} & \default{\textbf{72.7}} \\
    \text{[$x, y, w, h$]} & 74.5 & 84.2 & 70.2 \\
    [$x, y, w, h$]\dag & 75.4 & 85.0 & 72.2 \\
    [$(x,y)_{t-l}$$,$ $(x,y)_{b-r}$] & 75.2 & 84.4 & 71.0 \\
    \bottomrule
  \end{NiceTabular}
  \caption{\textbf{Sequence format study}.}
  \label{tab:seq}
  \vspace{-6pt}
\end{table}
Before where to look at the target, in which form (sequence format) to ``read out" the target's position is still controversial within generative paradigm trackers.
As \cref{tab:seq} presented, we explore three sequence formats: 1)  [$x_\text{min}$$,$ $y_\text{min}$$,$ $x_\text{max}$$,$ $y_\text{max}$] leverages a set of top-left and bottom-right corner coordinates to denote target's localization. It utilizes a shared vocabulary to depict the token corresponding to the coordinate, akin to the approach adopted by ARTrack~\cite{ARTrack}.
2) [$x, y, w, h$] denotes that generates the object's center point [$x, y$] and scale [$w, h$], employing a unified vocabulary that represents both position and height/width, as seen in the methodology by SeqTrack~\cite{SeqTrack}. Moreover, we experiment with decoupling the position and scale using separate vocabularies, denoted by \dag.
3) Drawing inspiration from Polyformer~\cite{liu2023polyformer}, we configure the sequence as [$(x,y)\text{t-l}$, $(x,y)\text{b-r}$], wherein ``t-l" represents top-left and ``b-r" represents bottom-right. This format includes two vertex coordinates, employing a 2D coordinate codebook. This codebook maps each floating-point coordinate to a bilinear interpolation of adjacent coordinates in 2D space without any quantization.

Our findings indicate that the [$x_\text{min}$, $y_\text{min}$, $x_\text{max}$, $y_\text{max}$] format surpasses the others. This observation primarily stems from 1) A unified vocabulary, contrary to our default setting, induces confusion between position and scale, resulting in lower precision 2) The variant [$x, y, w, h$]\dag, which decouples the position and scale using separate vocabularies, demonstrates better performance compared to a unified vocabulary. However, the utilization of multiple vocabularies introduces the potential interference between them. 3) [$(x,y)_\text{t-l}$$,$ $(x,y)_\text{b-r}$], which implies the utilization of almost infinite large vocabulary, significantly slows down the training and impaired accuracy.

\section{Trajectory and Appearance Evolution}

\def\arraystretch{1.0}
\renewcommand{\tabcolsep}{4 pt}
\begin{table}[h]\footnotesize
  \centering
  \begin{NiceTabular}{l|cccc}
    \toprule
    & AO(\%) & $SR_{0.5}$(\%) & $SR_{0.75}$(\%) & FPS\\
    \midrule
    \textbf{$\text{ARTrackV2}_\text{256}$} &  \default{\textbf{75.9}} & \default{\textbf{85.4}} & \default{\textbf{72.7}} & \default{94}\\
    w/o trajectory & 72.9 & 82.7 & 71.4 & 96 \\
    w/o appearance & 72.3 & 81.9 & 70.3 & 116 \\
    w/o both & 69.8 & 79.5 & 67.6 & \textbf{120}\\
    \bottomrule
  \end{NiceTabular}
  \caption{\textbf{Trajectory-appearance evolution}.}
  \label{tab:evo}
  \vspace{-6pt}
\end{table}

The primary focus of our work lies in the evolution of both trajectory and appearance. Therefore, it becomes imperative to conduct comparative experiments to assess the individual impacts of respective. Table \ref{tab:evo} illustrates the performance of these experiments, revealing that the exclusion of either trajectory or appearance evolution leads to a notable decline in performance.
This performance degradation is attributed to the fact that individual evolution components are insufficient to describe coherent target variation, particularly in intricate tracking scenarios. This underscores the significance of jointly considering trajectory and appearance evolution for optimal tracking performance.
Furthermore, our experiments involved the elimination of both trajectory and appearance evolution, resulting in a more pronounced decline in accuracy. In this variation, continuous tracking is transformed into a frame-level template-matching approach which breaks the time continue of tracking. 

\section{Masking Ratio for Appearance Modeling}
\def\arraystretch{1.0}
\renewcommand{\tabcolsep}{4 pt}
\begin{table}[h]\footnotesize
  \centering
\begin{NiceTabular}{c|ccc}
    \toprule
    ratio [\%] & AO & $SR_{0.5}$ & $SR_{0.75}$ \\
    \midrule
    0 & 74.7 & 84.0 & 70.8 \\
    25 & 75.0 & 84.2 & 70.9 \\
    50 & 75.2 & 84.5 & 71.2 \\
    75 & 75.5 & 84.8 & 71.9 \\
    \textbf{90} & \default{\textbf{75.9}} & \default{\textbf{85.4}} & \default{\textbf{72.7}} \\
    95 & 75.3 & 84.7 & 71.6 \\
    \bottomrule
\end{NiceTabular}
  \caption{\textbf{Masking ratio study}.}
  \label{tab:mask}
  \vspace{0pt}
\end{table}
In ARTrackV2, a key aspect of the design is the incorporation of an exceptionally high masking ratio. During the training phase, we randomly mask appearance tokens after feeding them into the transformer encoder based on a predetermined masking ratio prior to target reconstruction. The influence of various masking ratios on model accuracy is presented in Table \ref{tab:mask}. Unlike BERT\cite{bert} and MAE\cite{mae}, our findings indicate that the model accuracy consistently improves as the masking rate increases until reaching a masking ratio of 90\%. We posit that this phenomenon may arise from the redundancy and relevance of temporal information in comparison to textual and image cues\cite{tong2022videomae, videomaev2}. Consequently, even when subjected to exceptionally high occlusion ratios, \text{ARTrackV2} exhibits the capability to generate rational outputs.

\def\fwidth{0.18\linewidth}
\def\arraystretch{0.5}
\renewcommand{\tabcolsep}{0.5 pt}
\begin{figure}[t]
\centering
\begin{tabular}{cccccc}
    \multirow{2}{*}{\rotatebox{90}{\scriptsize{Variation}}} \hspace{1pt} &
    \includegraphics[width=\fwidth]{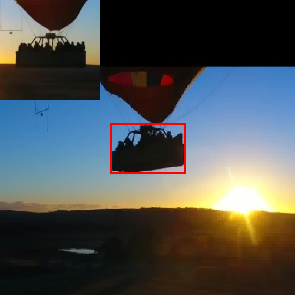} & 
     \includegraphics[width=\fwidth]{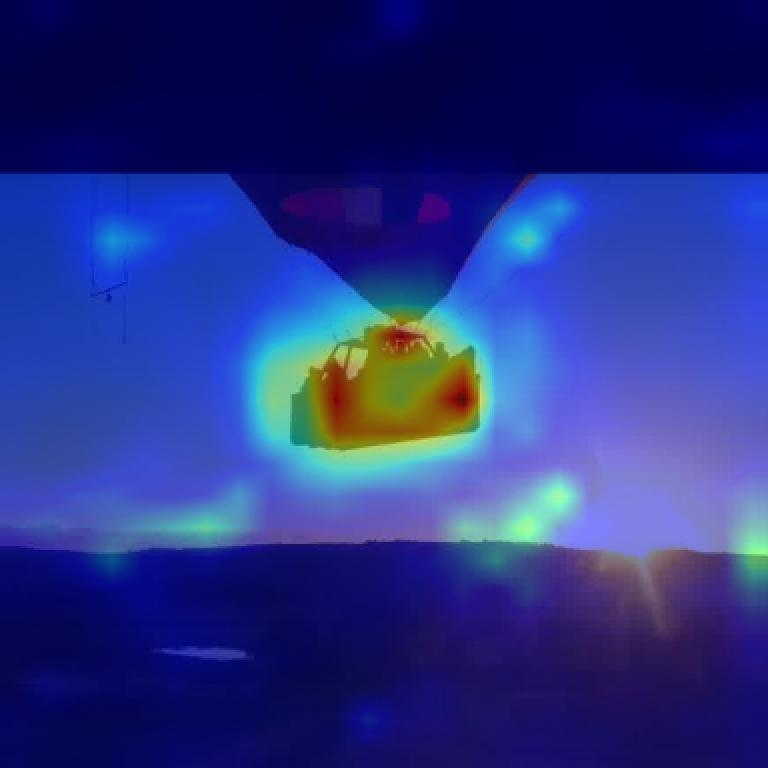} &  
     \includegraphics[width=\fwidth]{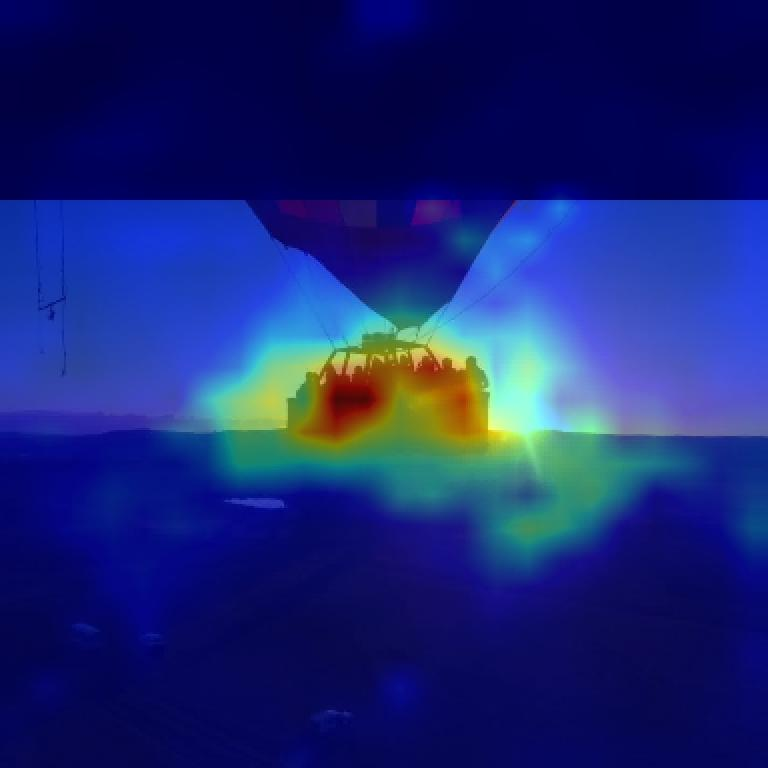} &
     \includegraphics[width=\fwidth]{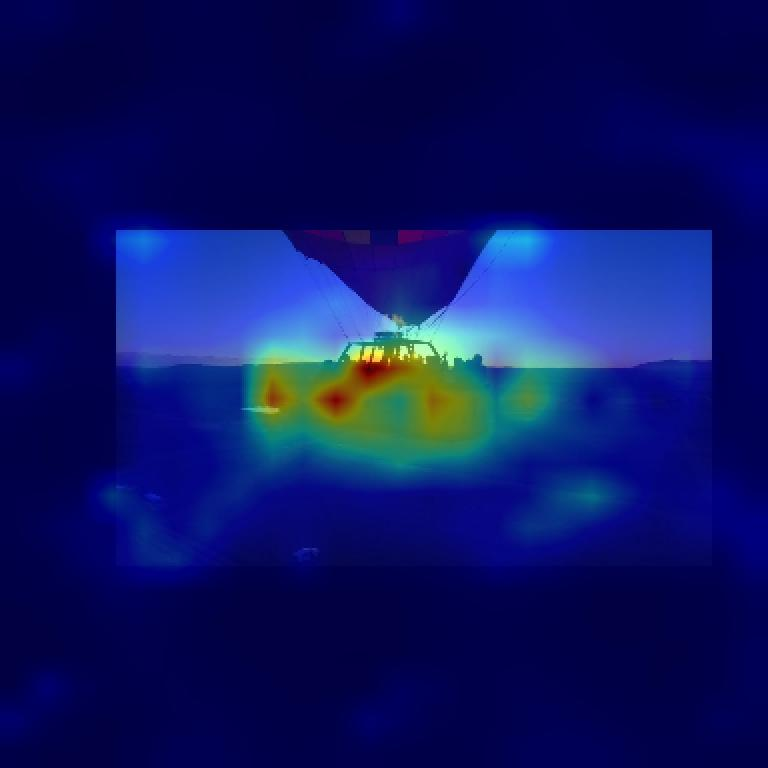} & 
     \includegraphics[width=\fwidth]{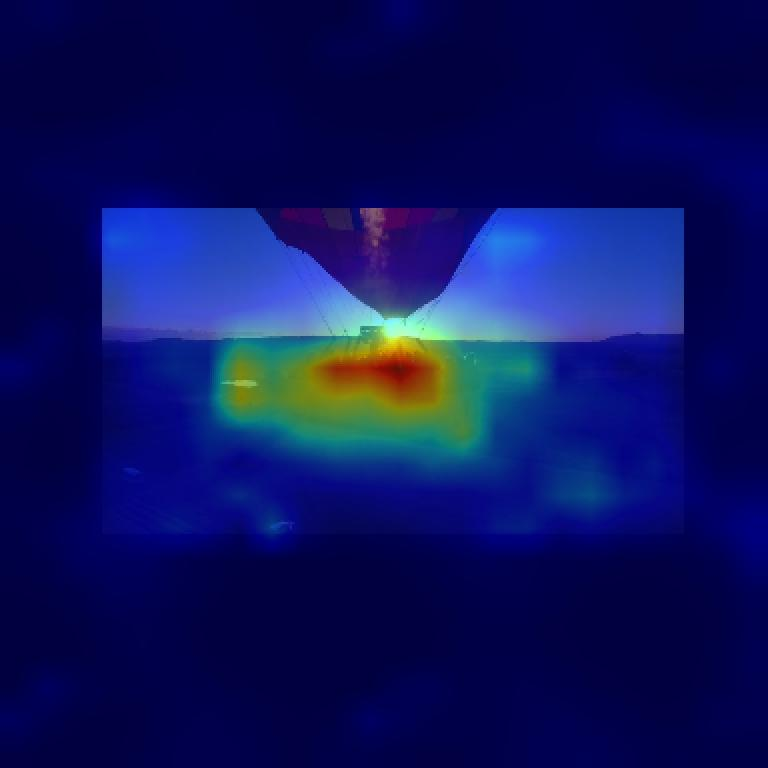} \\
     & \includegraphics[width=\fwidth]{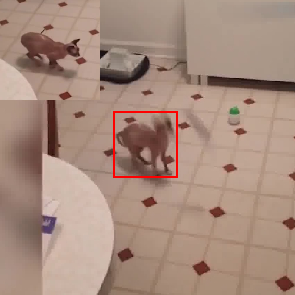} 
     & \includegraphics[width=\fwidth]{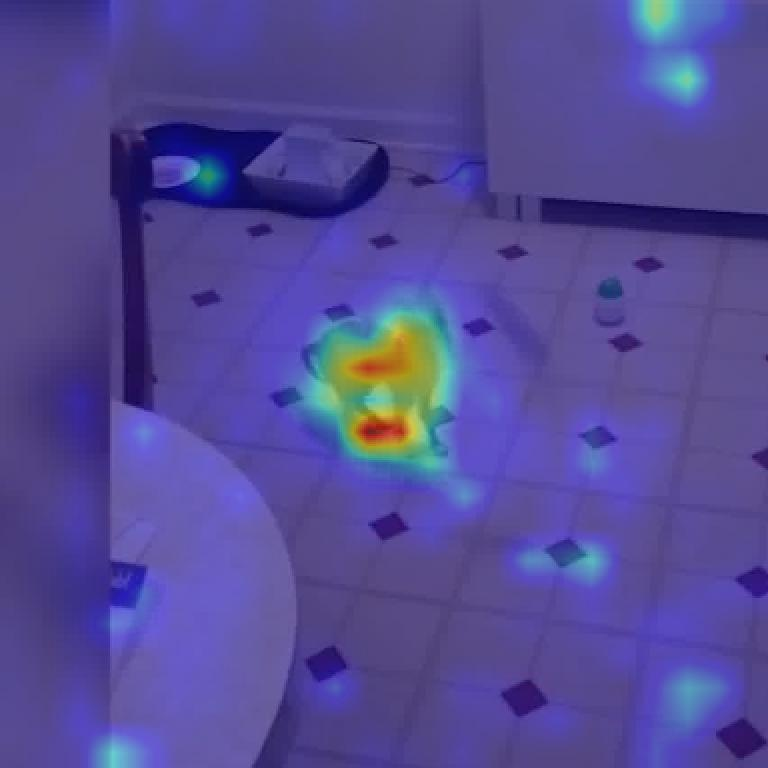} &  
     \includegraphics[width=\fwidth]{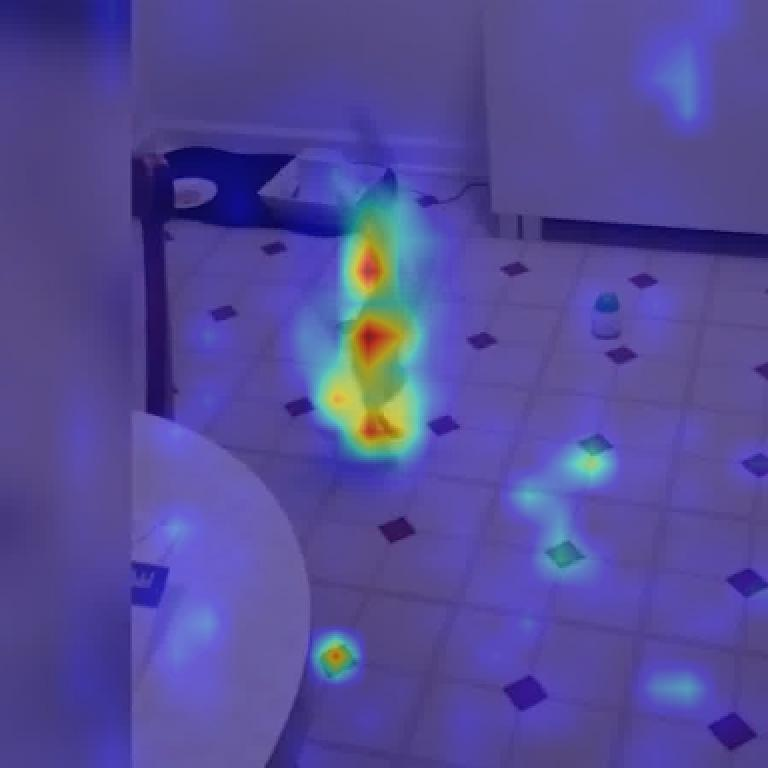} &
     \includegraphics[width=\fwidth]{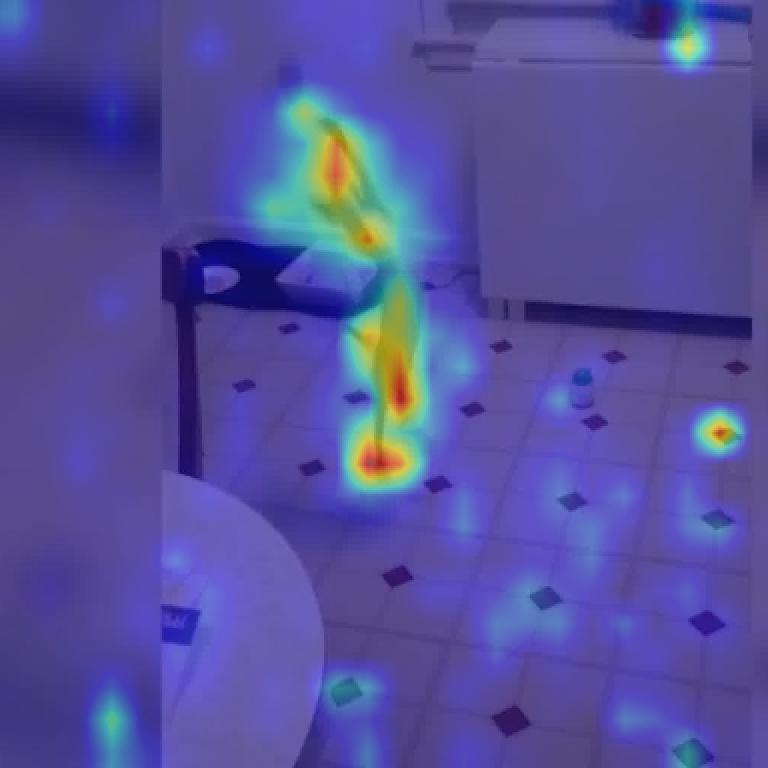} & 
     \includegraphics[width=\fwidth]{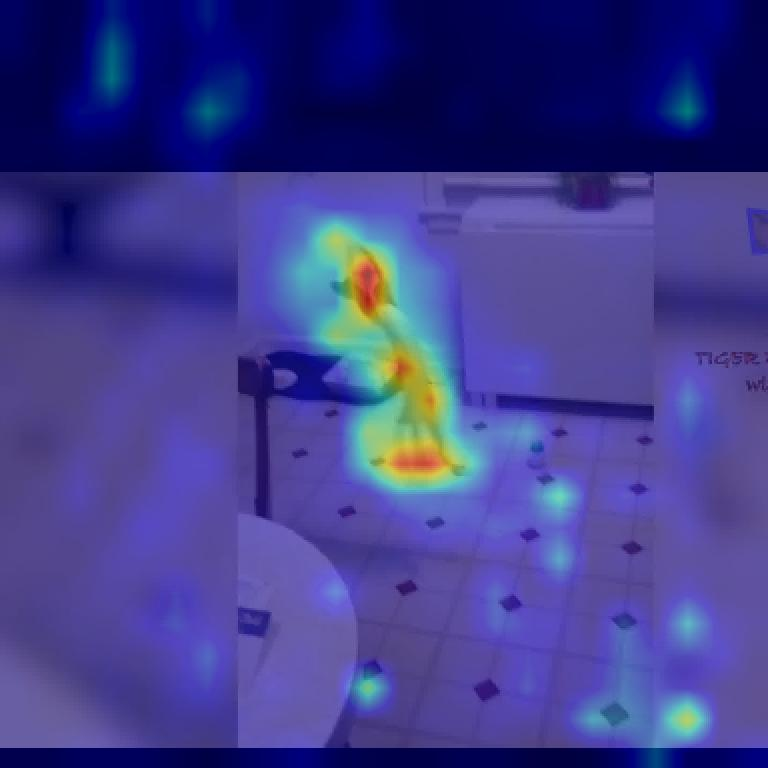} \\
     \multirow{2}{*}{\rotatebox{90}{\scriptsize{Occlusion}}} \hspace{1pt} &
     \includegraphics[width=\fwidth]{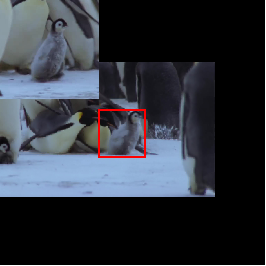} &
     \includegraphics[width=\fwidth]{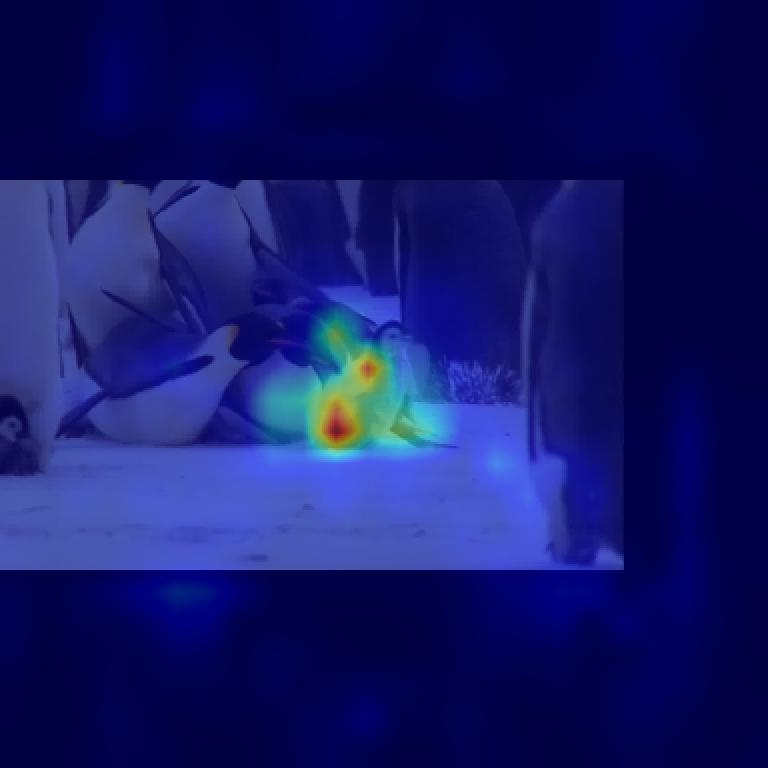} &  
     \includegraphics[width=\fwidth]{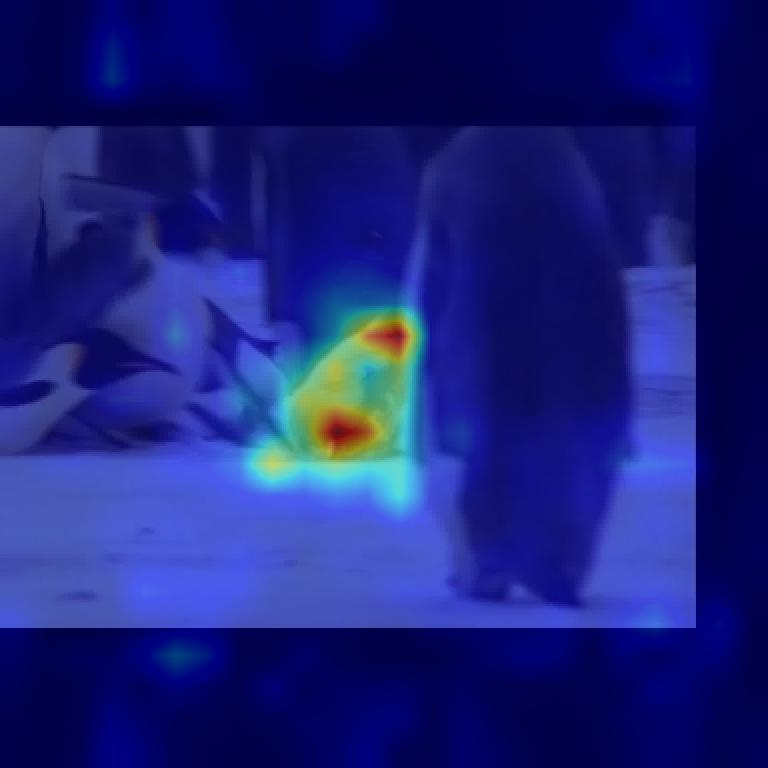} &
     \includegraphics[width=\fwidth]{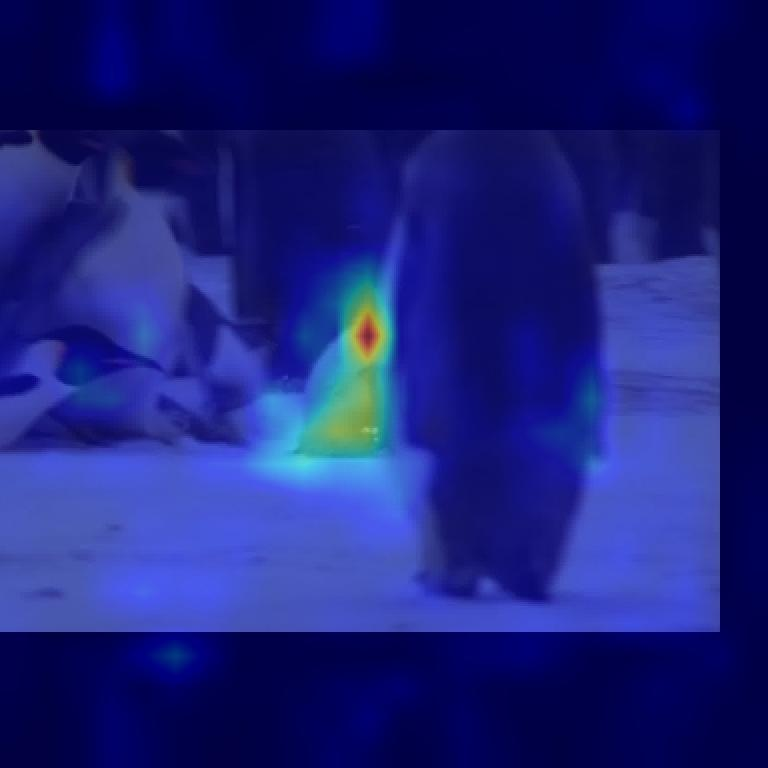} & 
     \includegraphics[width=\fwidth]{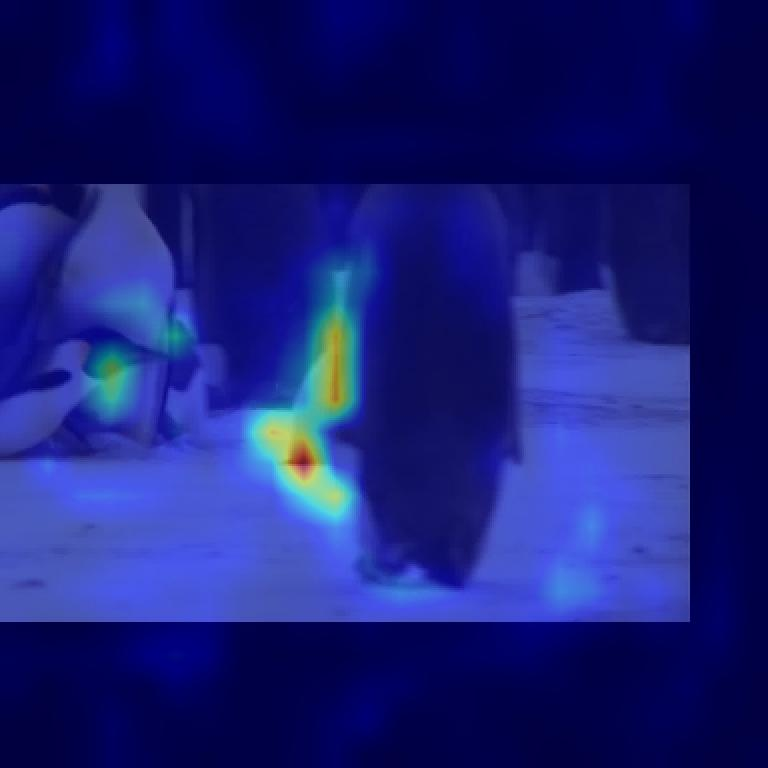} \\
     & \includegraphics[width=\fwidth]{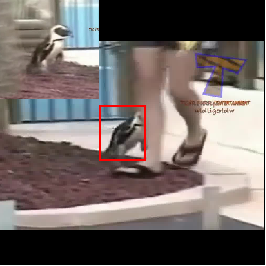} &
     \includegraphics[width=\fwidth]{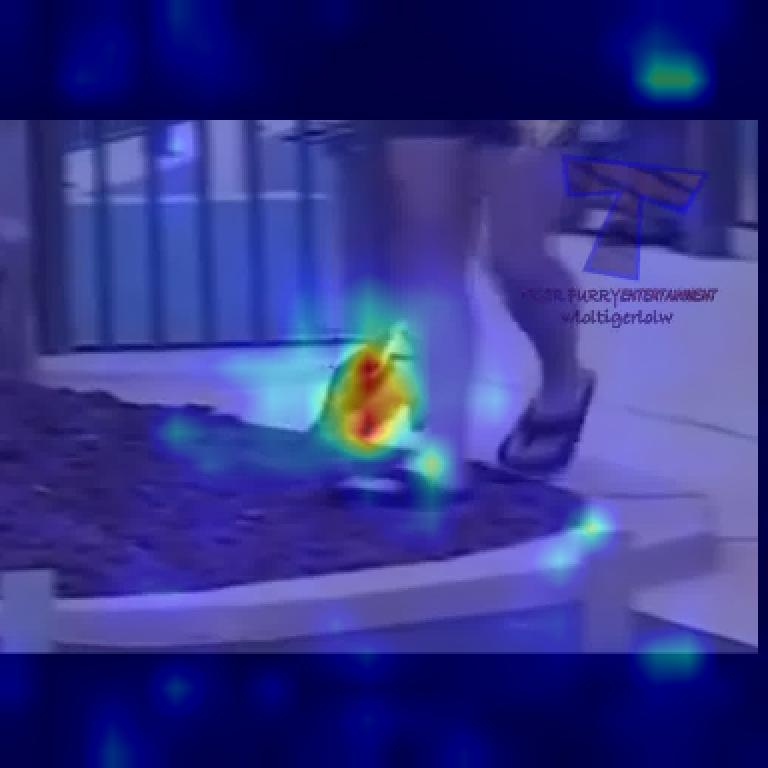} &  
     \includegraphics[width=\fwidth]{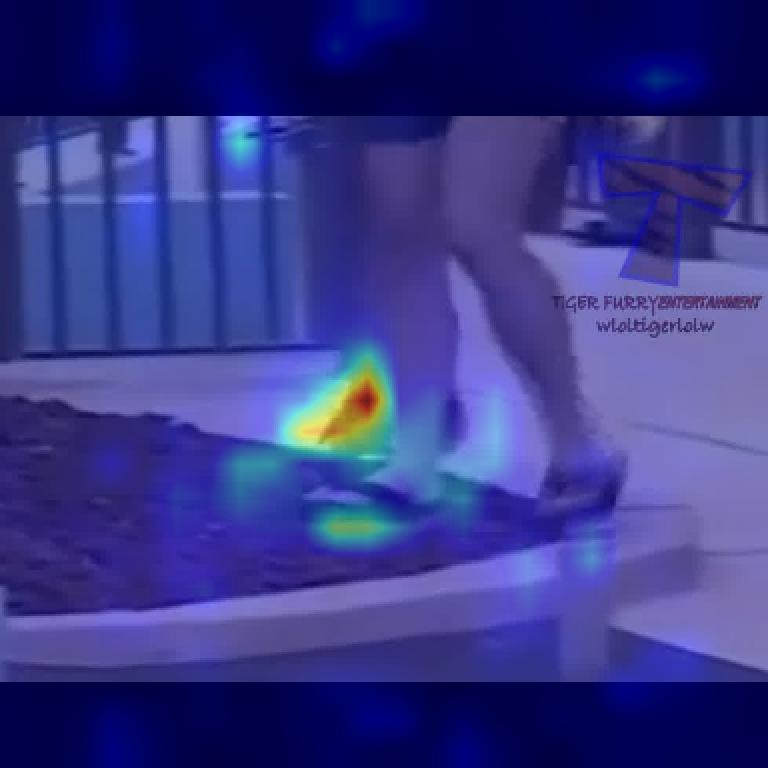} &
     \includegraphics[width=\fwidth]{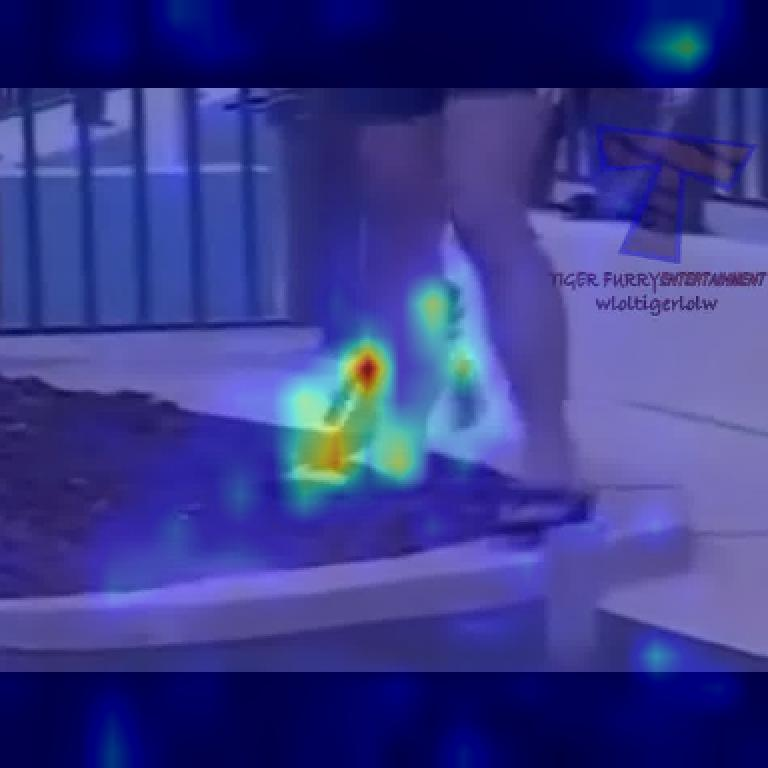} & 
     \includegraphics[width=\fwidth]{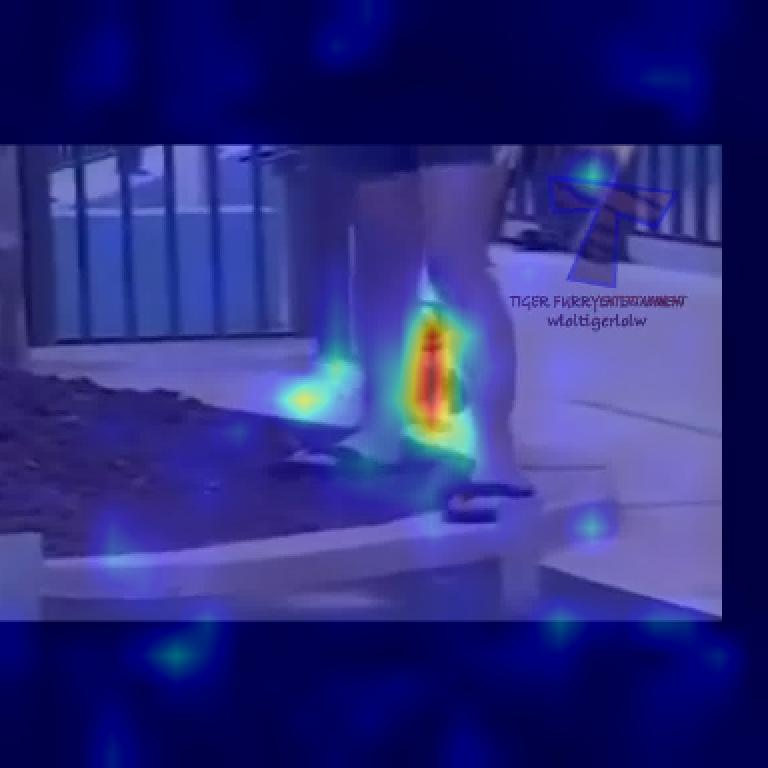} \\
         \multirow{2}{*}{\rotatebox{90}{\scriptsize{Distractor}}} \hspace{1pt} &
    \includegraphics[width=\fwidth]{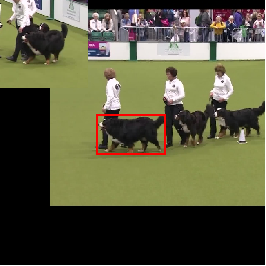} & 
     \includegraphics[width=\fwidth]{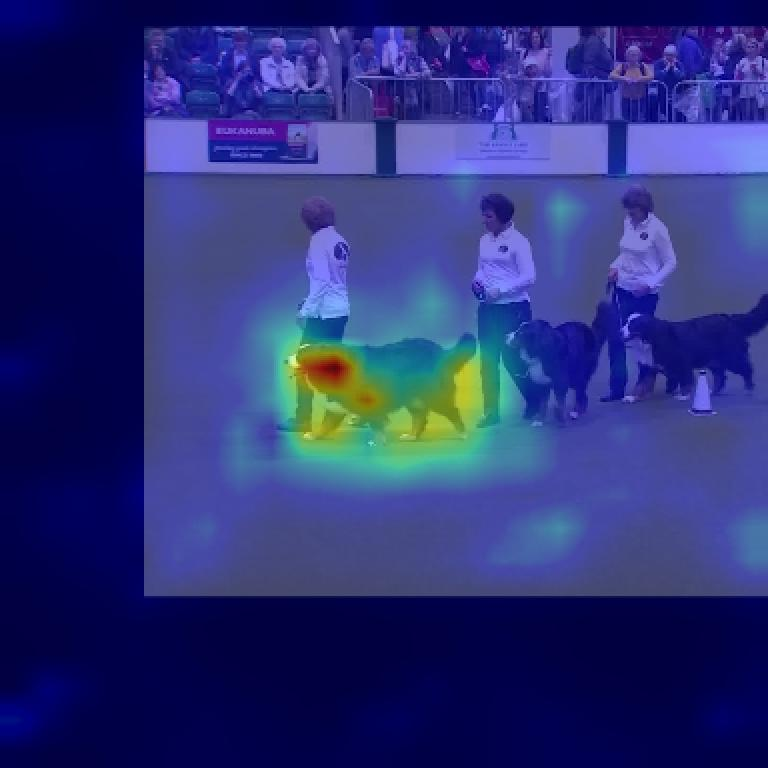} &  
     \includegraphics[width=\fwidth]{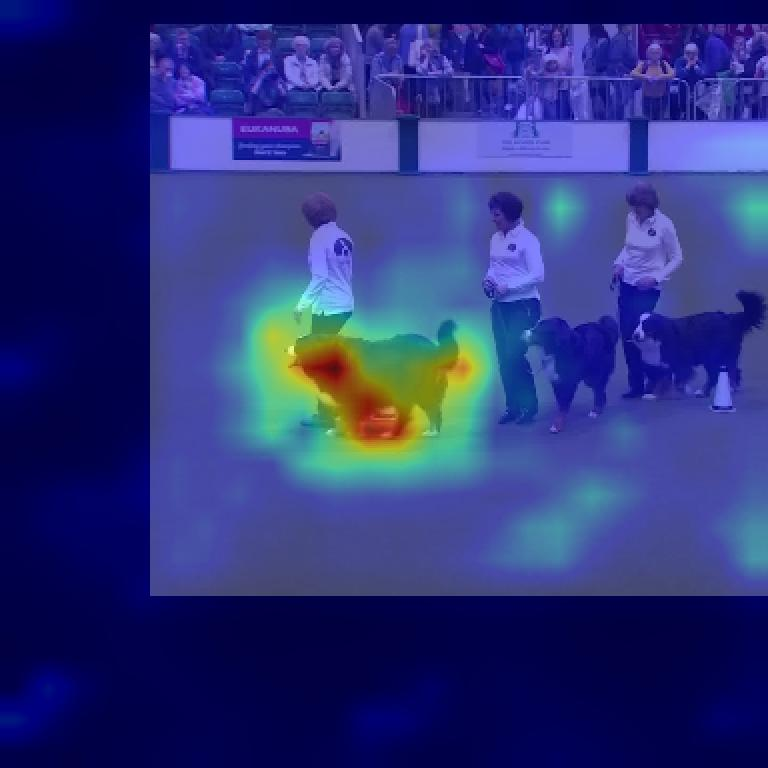} &
     \includegraphics[width=\fwidth]{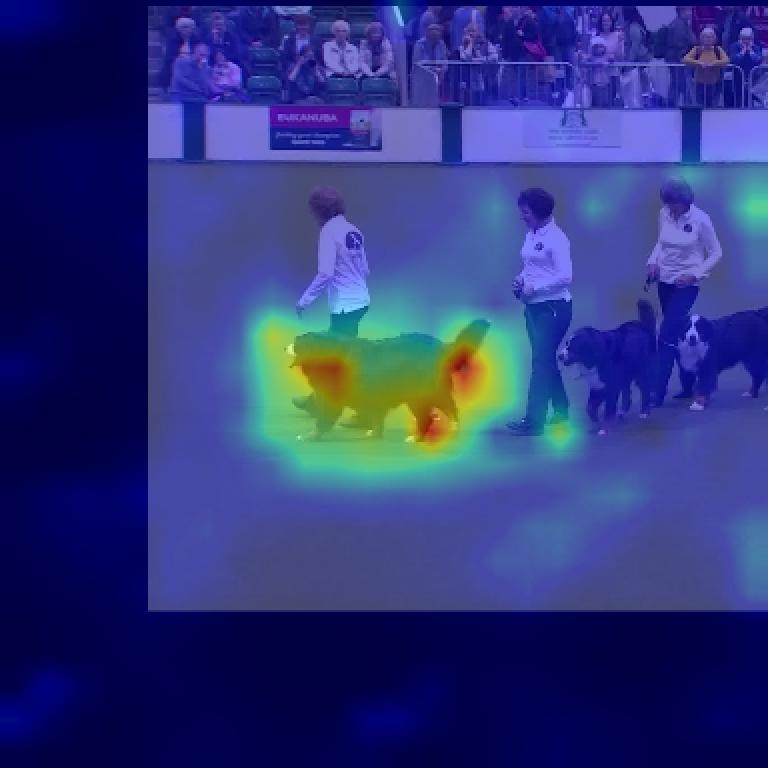} & 
     \includegraphics[width=\fwidth]{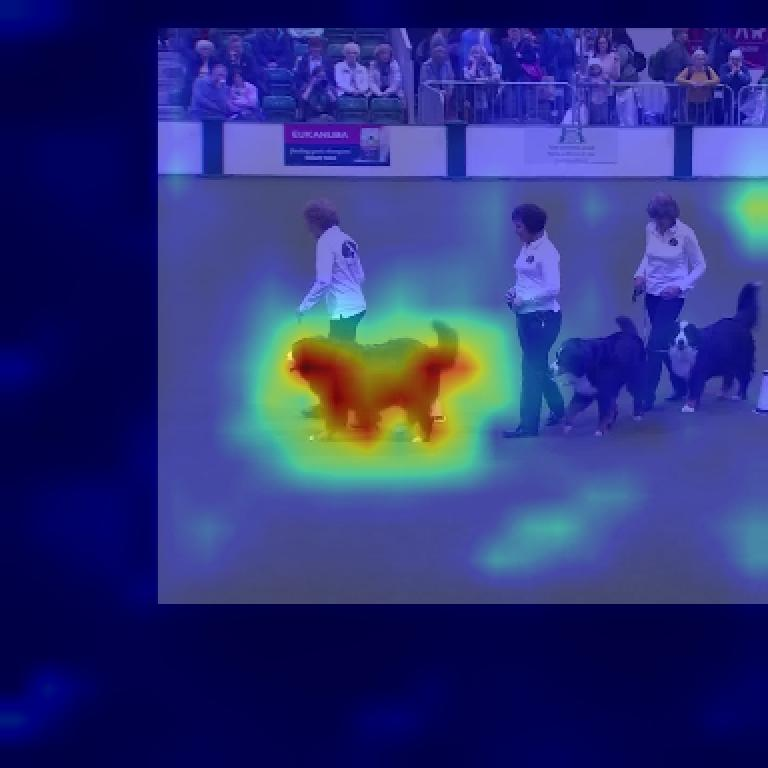} \\
     & \includegraphics[width=\fwidth]{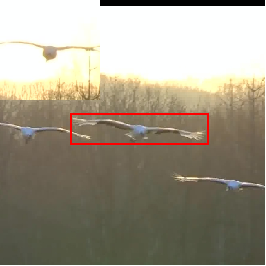} 
     & \includegraphics[width=\fwidth]{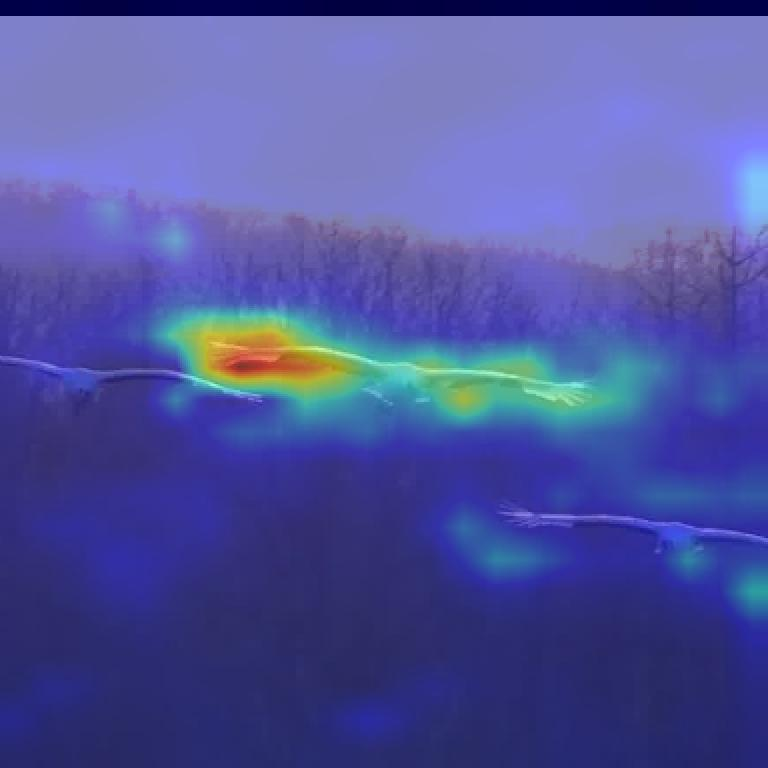} &  
     \includegraphics[width=\fwidth]{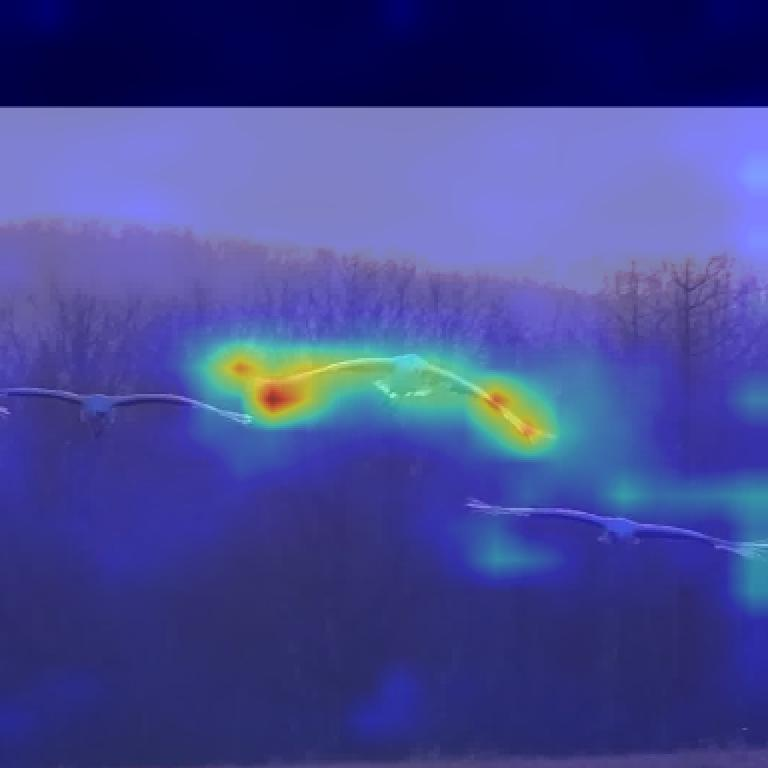} &
     \includegraphics[width=\fwidth]{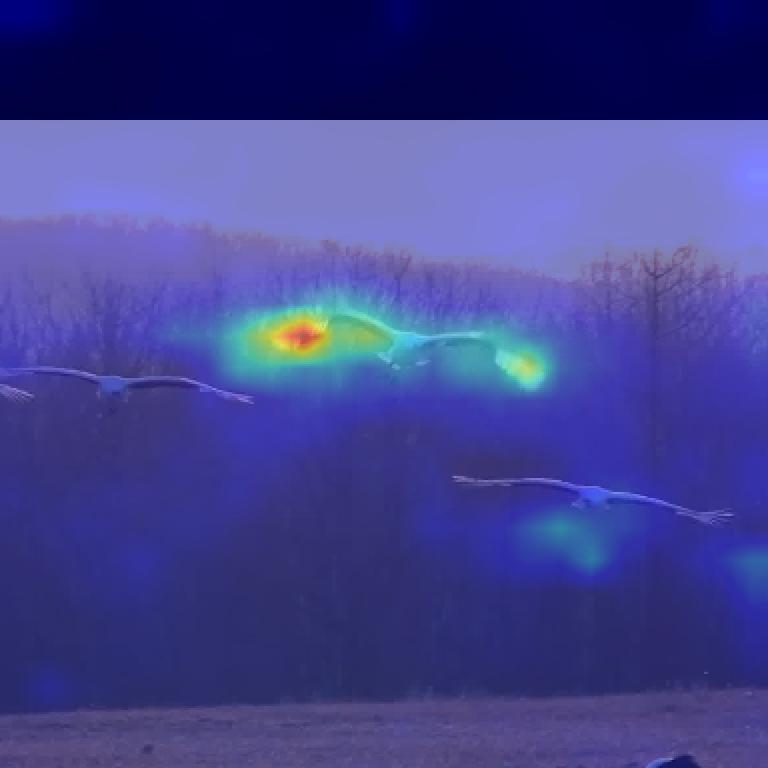} & 
     \includegraphics[width=\fwidth]{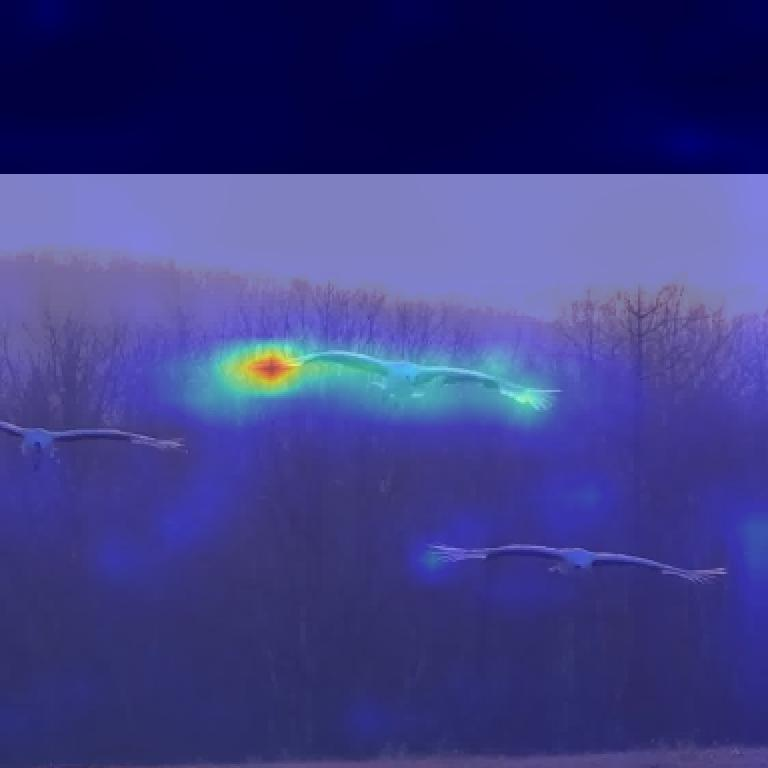} \\
     \multirow{2}{*}{\rotatebox{90}{\scriptsize{Motion Blur}}} \hspace{1pt} &
     \includegraphics[width=\fwidth]{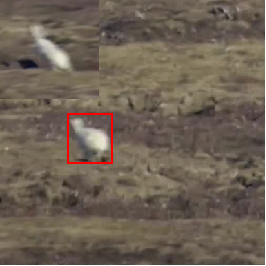} &
     \includegraphics[width=\fwidth]{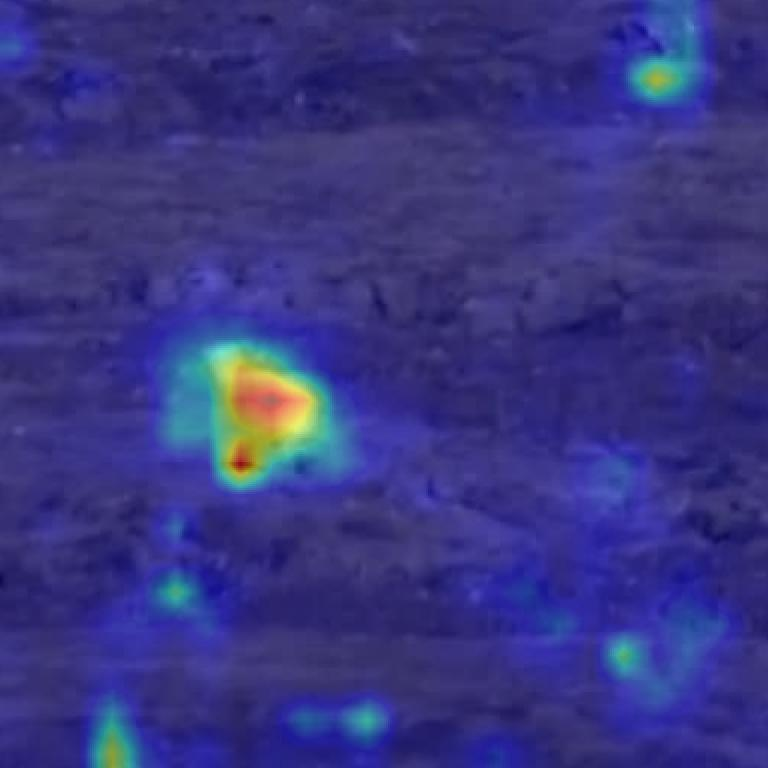} &  
     \includegraphics[width=\fwidth]{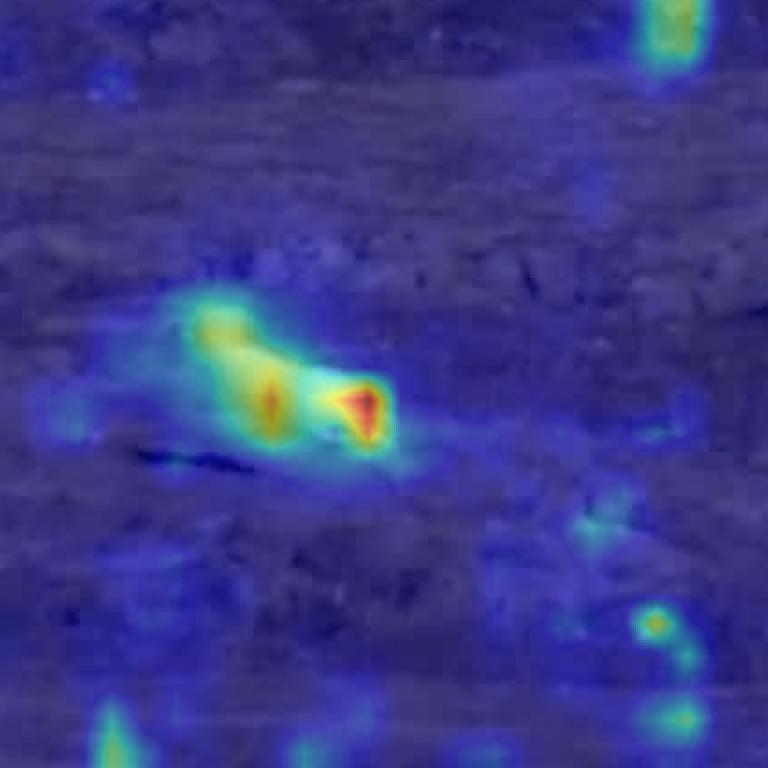} &
     \includegraphics[width=\fwidth]{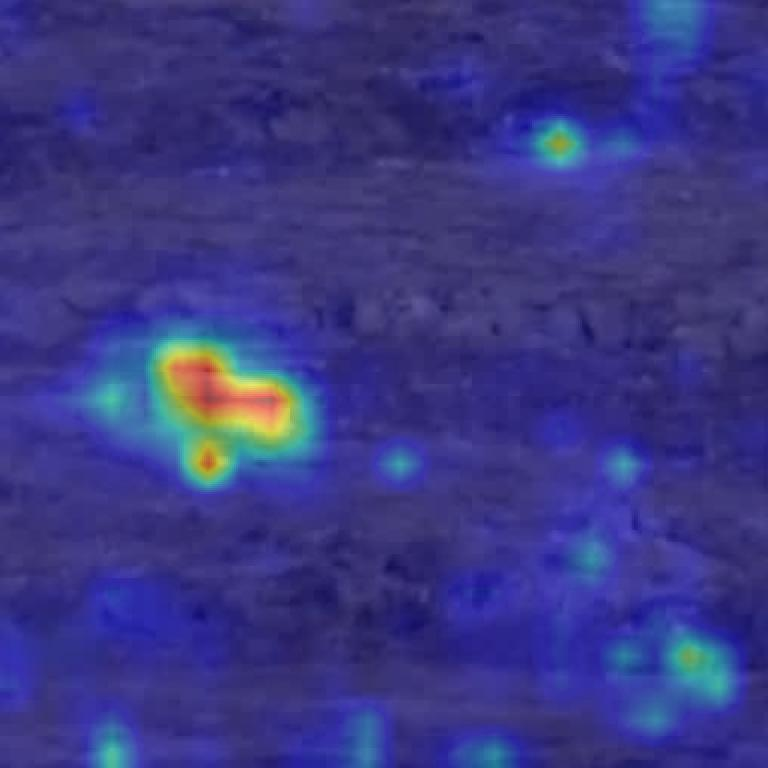} & 
     \includegraphics[width=\fwidth]{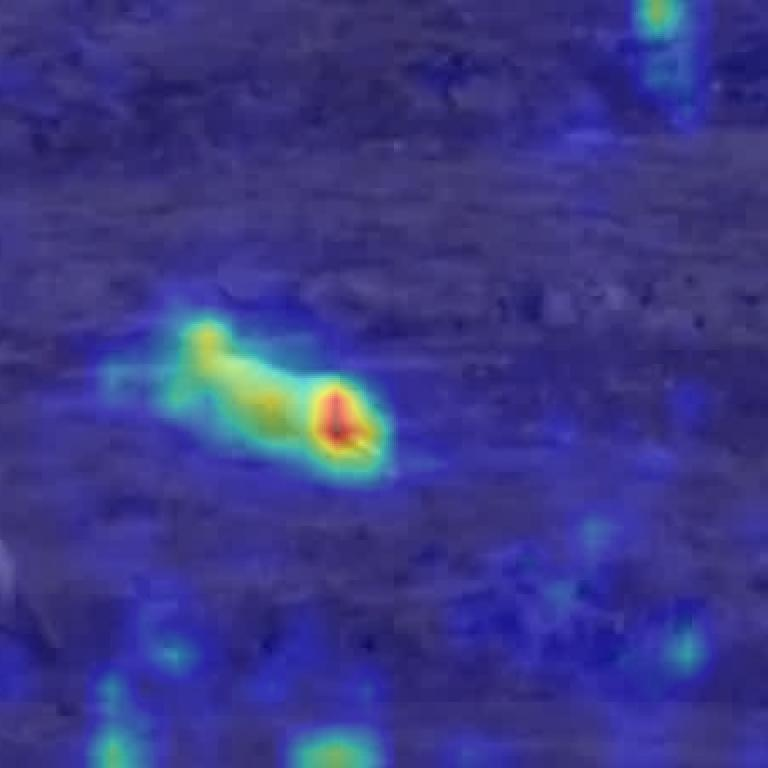} \\
     & \includegraphics[width=\fwidth]{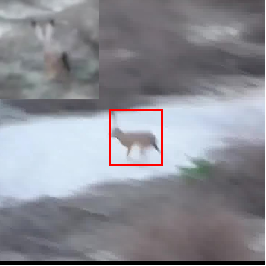} &
     \includegraphics[width=\fwidth]{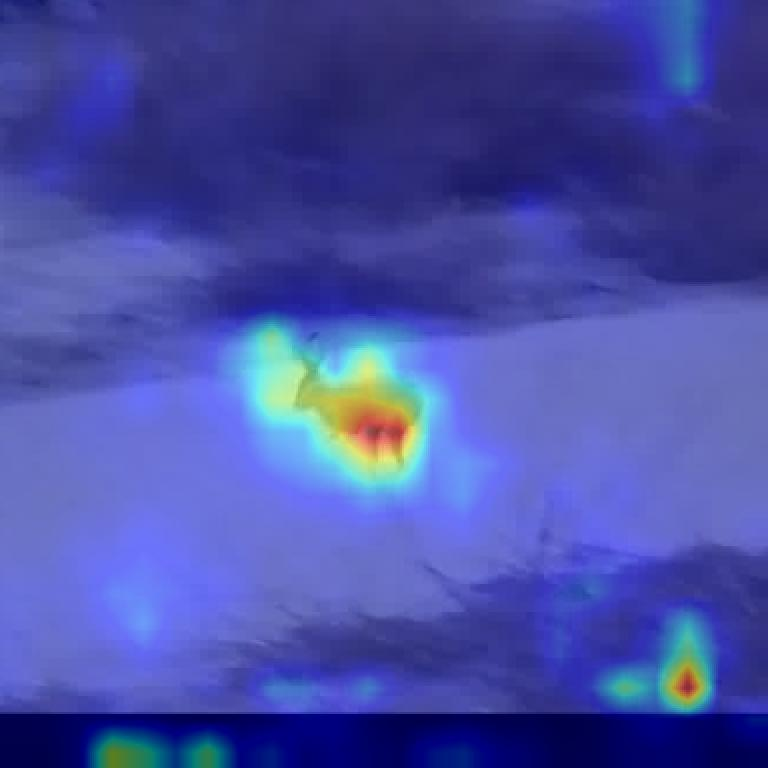} &  
     \includegraphics[width=\fwidth]{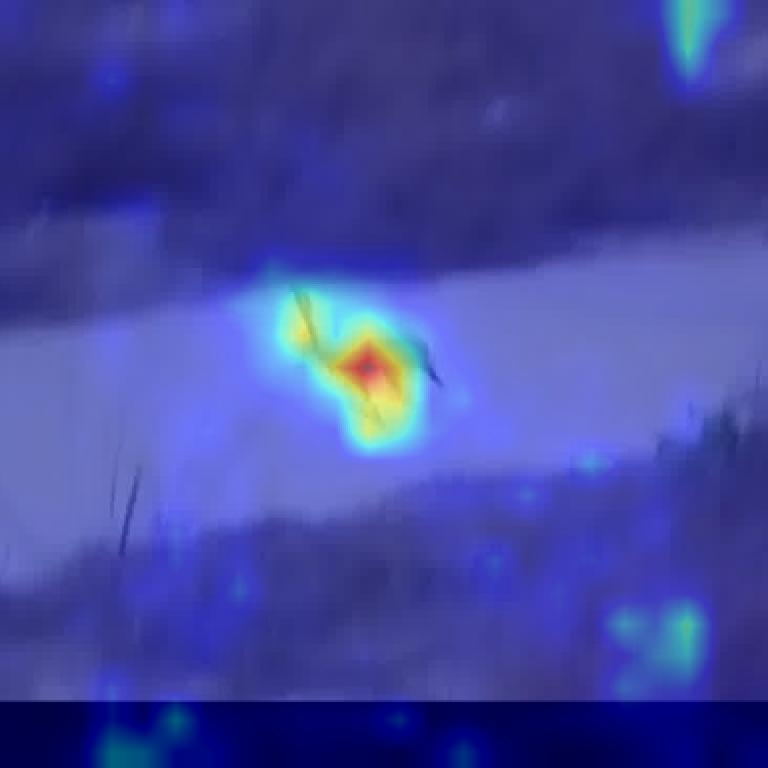} &
     \includegraphics[width=\fwidth]{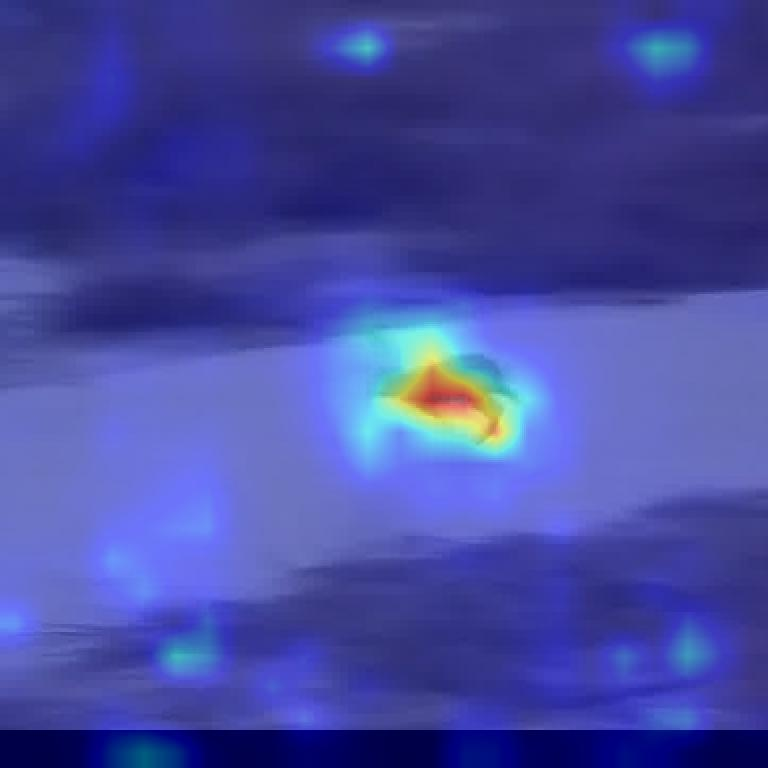} & 
     \includegraphics[width=\fwidth]{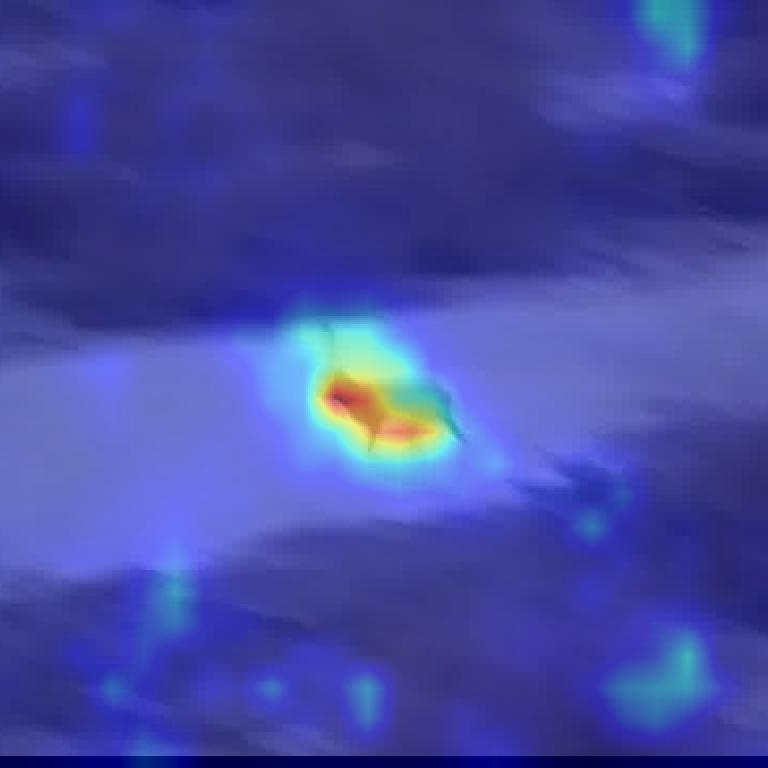} \\
     & \scriptsize{(a) search} & \scriptsize{(b) time $t$} & \scriptsize{(c) time $t+1$} & \scriptsize{(d) time $t+2$} & \scriptsize{(e) time $t+3$} \\ 
\end{tabular}
\caption{\textbf{More attention visualization}. (a): Search region and template. The \textcolor{red}{red} boxes denote the ground truth. (b)-(e): Appearance tokens to search the cross-attention map of ARTrackV2.}
\label{fig:more_visual}
\end{figure}

\section{More Visualization of Appearance Tokens}

Within this section, we introduce additional visualizations of the cross-attention map between appearance tokens and the search region. As depicted in Figure \ref{fig:more_visual}, the findings from these visual representations align with the previous spot: the visualizations substantiate the adaptability and versatility of our model, particularly evident in challenging scenarios.

Additionally, our endeavor to visualize appearance tokens in an intuitive way encountered considerable challenges. We utilized dimensionality reduction techniques like t-distributed stochastic neighbor embedding (t-SNE)~\cite{van2008visualizing} to transform high-dimensional appearance tokens into one-dimensional spaces, aiming to produce grayscale images. Regrettably, the resolution of appearance tokens within the feature space remains notably limited. Consequently, the generated grayscale images are blurry with unclear boundaries due to this low-resolution representation.

\def\fwidth{0.18\linewidth}
\def\arraystretch{0.5}
\renewcommand{\tabcolsep}{0.5 pt}
\begin{figure}[t]
\centering
\begin{tabular}{cccccc}
    \multirow{2}{*}{\rotatebox{90}{\scriptsize{Variation}}} \hspace{1pt} &
    \includegraphics[width=\fwidth]{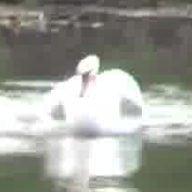} & 
     \includegraphics[width=\fwidth]{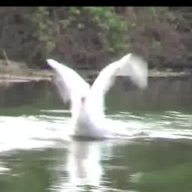} &  
     \includegraphics[width=\fwidth]{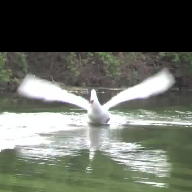} &
     \includegraphics[width=\fwidth]{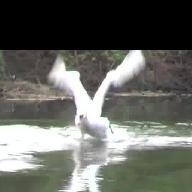} & 
     \includegraphics[width=\fwidth]{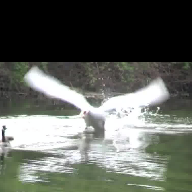} \\
     & \includegraphics[width=\fwidth]{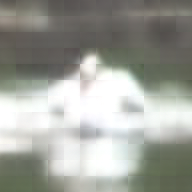} 
     & \includegraphics[width=\fwidth]{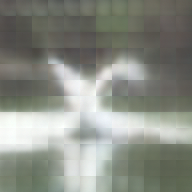} &  
     \includegraphics[width=\fwidth]{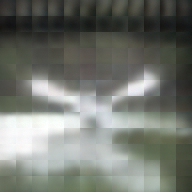} &
     \includegraphics[width=\fwidth]{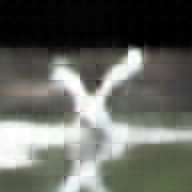} & 
     \includegraphics[width=\fwidth]{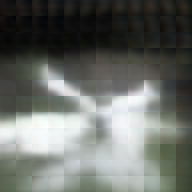} \\
     \multirow{2}{*}{\rotatebox{90}{\scriptsize{Occlusion}}} \hspace{1pt} &
     \includegraphics[width=\fwidth]{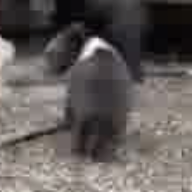} &
     \includegraphics[width=\fwidth]{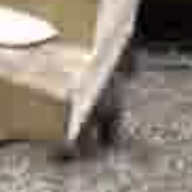} &  
     \includegraphics[width=\fwidth]{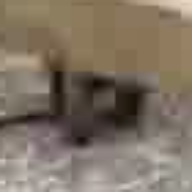} &
     \includegraphics[width=\fwidth]{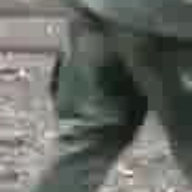} & 
     \includegraphics[width=\fwidth]{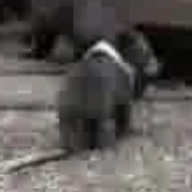} \\
     & \includegraphics[width=\fwidth]{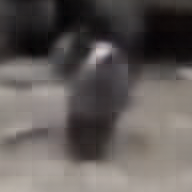} &
     \includegraphics[width=\fwidth]{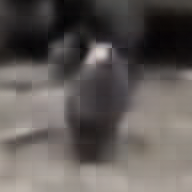} &  
     \includegraphics[width=\fwidth]{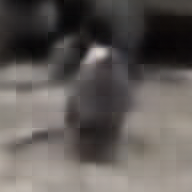} &
     \includegraphics[width=\fwidth]{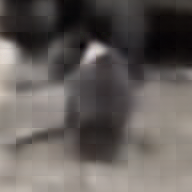} & 
     \includegraphics[width=\fwidth]{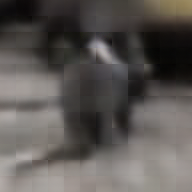} \\
          \multirow{2}{*}{\rotatebox{90}{\scriptsize{Out-of-view}}} \hspace{1pt} &
     \includegraphics[width=\fwidth]{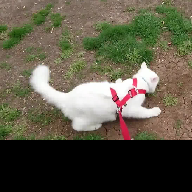} &
     \includegraphics[width=\fwidth]{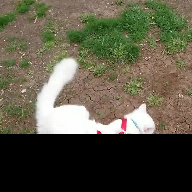} &  
     \includegraphics[width=\fwidth]{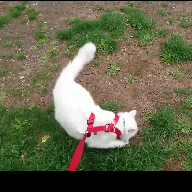} &
     \includegraphics[width=\fwidth]{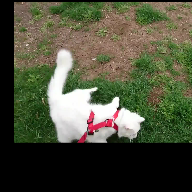} & 
     \includegraphics[width=\fwidth]{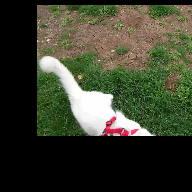} \\
     & \includegraphics[width=\fwidth]{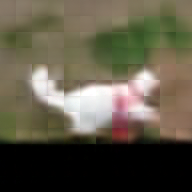} &
     \includegraphics[width=\fwidth]{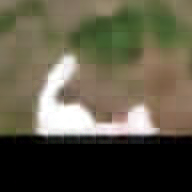} &  
     \includegraphics[width=\fwidth]{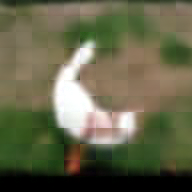} &
     \includegraphics[width=\fwidth]{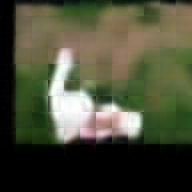} & 
     \includegraphics[width=\fwidth]{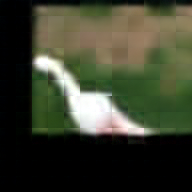} \\
          \multirow{2}{*}{\rotatebox{90}{\scriptsize{ 
           Dimming}}} \hspace{1pt} &
     \includegraphics[width=\fwidth]{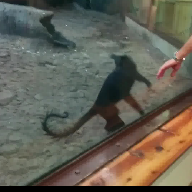} &
     \includegraphics[width=\fwidth]{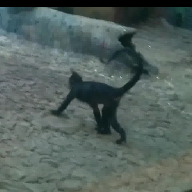} &  
     \includegraphics[width=\fwidth]{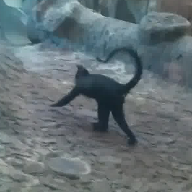} &
     \includegraphics[width=\fwidth]{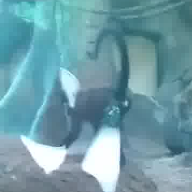} & 
     \includegraphics[width=\fwidth]{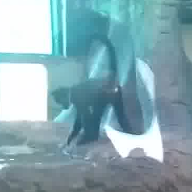} \\
     & \includegraphics[width=\fwidth]{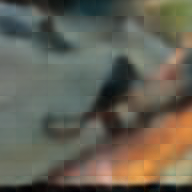} &
     \includegraphics[width=\fwidth]{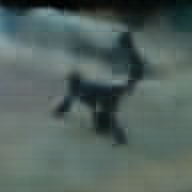} &  
     \includegraphics[width=\fwidth]{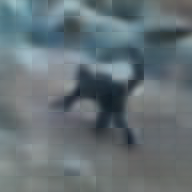} &
     \includegraphics[width=\fwidth]{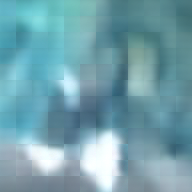} & 
     \includegraphics[width=\fwidth]{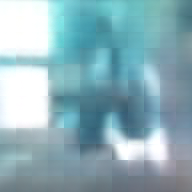} \\
     & \scriptsize{(a) time $t$} & \scriptsize{(b) time $t+N$} & \scriptsize{(c) time $t+2N$} & \scriptsize{(d) time $t+3N$} &
     \scriptsize{(e) time $t+4N$}\\ 
\end{tabular}
\caption{
\textbf{Image reconstruction visualizations}. Reconstructed appearance in the image pixel domain of GOT-10k val set. For each video sequence, we show the target's appearance (top) and reconstructed appearance image (bottom). (a)-(e) represented sampling video frames at a fixed interval of N.
}
\label{fig:image}
\end{figure} 

\section{Visualization of Image Reconstruction}

To offer a more intuitive portrayal of our appearance evolution, distinct from visualizing the cross-attention map within the feature domain between appearance tokens and search, we conduct a quantitative analysis of appearance reconstruction within the image pixel domain. This analysis leverages an image reconstruction objective, aimed at reconstructing the target within the image pixel domain, as detailed in the ablation experiment.
 Illustrated in Figure \ref{fig:image}, the first two rows depict tracking under appearance variation scenarios. ARTrackV2 adeptly captures long-term appearance evolution, ensuring accurate appearance descriptions even amidst significant changes, thereby facilitating precise tracking in complex environments.
The subsequent two rows showcase tracking in occlusion scenarios. It's evident that when the target is unobstructed, ARTrackV2 accurately captures appearance changes. However, in instances where the target becomes occluded, the appearance token reconstructs the previous visible appearance. This prevents erroneous propagation of appearance, which could otherwise mislead the model by providing nonsensical target localization. Upon the reappearance of the target, its appearance is accurately captured and faithfully reconstructed.
The following two rows exemplify our tracker's proficiency in managing out-of-view targets. As the object gradually moves out of sight, our tracker adeptly adjusts to appearance changes, ensuring a seamless evolution in the reconstructed target's appearance.
In the final two rows, the target's illumination gradually fluctuates. The appearance model notably captures these illumination variations with precision, resulting in a smoother and clearer evolution in appearance.

%% file: figs/tnl2k_nfs_uav123.tex
\pgfplotsset{
    bar group size/.style 2 args={
        /pgf/bar shift={%
                -0.5*(#2*\pgfplotbarwidth + (#2-1)*\pgfkeysvalueof{/pgfplots/bar group skip})  + 
                (.5+#1)*\pgfplotbarwidth + #1*\pgfkeysvalueof{/pgfplots/bar group skip}},%
    },
    bar group skip/.initial=1.0pt,
    plot 0/.style={color=black,fill=cSiamFC,fill,text=black},%
    plot 1/.style={color=black,fill=cECO,fill,text=black},%
    plot 2/.style={color=black,fill=cOcean,fill,text=black},%
    plot 3/.style={color=black,fill=cATOM,fill,text=black},%
    plot 4/.style={color=black,fill=cDiMP,fill,text=black},%
    plot 5/.style={color=black,fill=cTransT,fill,text=black},%
    plot 6/.style={color=black,fill=cSTARK,fill,text=black},%
    plot 7/.style={color=black,fill=cMixFormer-L,fill,text=black},%
    plot 8/.style={color=black,fill=cOSTrack,fill,text=black},%
    plot 9/.style={color=black,fill=cARTrack-256,fill,text=black,},%
    plot 10/.style={color=black,fill=cARTrack-384,fill,text=black},%
    plot 11/.style={color=black,fill=cARTrack-L-384,fill,text=black},%
    plot 14/.style={color=black,fill=cCTTrack,fill,text=black},%
    plot 15/.style={color=black,fill=cGRM,fill,text=black},%
    plot 16/.style={color=black,fill=cMixViT,fill,text=black},%
    plot 17/.style={color=black,fill=cSeqTrack,fill,text=black},%
    plot 12/.style={color=black,fill=cARTrackV2-256,fill,text=black,postaction={semithick,pattern = north west lines,pattern color = black}},%
    plot 13/.style={color=black,fill=cARTrackV2-L-384,fill,text=black,postaction={semithick,pattern = north east lines,pattern color = black}},%
}
\pgfplotsset{ every non boxed x axis/.append style={x axis line style=-},
     every non boxed y axis/.append style={y axis line style=-}}

\begin{figure*}
    \centering
    \begin{tikzpicture}[/pgfplots/width=1.04\linewidth,/pgfplots/height=0.24\linewidth,font=\scriptsize]
    \begin{axis}[axis x line = bottom,axis y line = none,
    ymin=40,ymax=72,
    ybar, bar width=12pt,
    enlargelimits=0.25,
    symbolic x coords={TNL2K,NFS,UAV123},
    xtick={TNL2K,NFS,UAV123},
	nodes near coords,
	nodes near coords style={font=\fontsize{4.5}{6}\selectfont},
	legend style={font=\fontsize{6}{7}\selectfont,draw=none,anchor=west,legend columns=-1,at={(0.0,1.12)},/tikz/every even column/.append style={column sep=0.08cm}},
	legend cell align={left},
    ]
    \addplot[plot 3,bar group size={0}{12}] coordinates {(UAV123,63.2)};
    \addplot[plot 4,bar group size={1}{12}] coordinates {(UAV123,64.3)};
    \addplot[plot 5,bar group size={2}{12}] coordinates {(UAV123,68.1)};
    \addplot[plot 8,bar group size={3}{12}] coordinates {(UAV123,70.7)};
    \addplot[plot 14,bar group size={4}{12}] coordinates {(UAV123,71.3)};
    \addplot[plot 15,bar group size={5}{12}] coordinates {(UAV123,70.2)};
    \addplot[plot 16,bar group size={6}{12}] coordinates {(UAV123,68.7)};
    \addplot[plot 17,bar group size={7}{12}] coordinates {(UAV123,68.5)};
    \addplot[plot 9,bar group size={8}{12}] coordinates {(UAV123,67.7)};
    \addplot[plot 12,bar group size={9}{12}] coordinates {(UAV123,69.9)};\label{bASTrackL}
    \addplot[plot 11,bar group size={10}{12}] coordinates {(UAV123,71.2)};\label{bASTrackL}
    \addplot[plot 13,bar group size={11}{12}] coordinates {(UAV123,71.7)};\label{bASTrackL}
    \legend{ATOM, DiMP, TransT, OSTrack, CTTrack, GRM, MixViT, SeqTrack, $\text{ARTrack}_\text{256}$,$\textbf{ARTrackV2}_\textbf{256}$,$\text{ARTrack-L}_ \text{384}$, $\textbf{ARTrackV2-L}_\textbf{384}$}
    
    \addplot[plot 3,bar group size={0}{10}] coordinates {(NFS,58.3)};
    \addplot[plot 4,bar group size={1}{10}] coordinates {(NFS,61.8)};
    \addplot[plot 5,bar group size={2}{10}] coordinates {(NFS,65.3)};
    \addplot[plot 8,bar group size={3}{10}] coordinates {(NFS,66.5)};
    \addplot[plot 15,bar group size={4}{10}] coordinates {(NFS,65.6)};
    \addplot[plot 17,bar group size={5}{10}] coordinates {(NFS,66.2)};
    \addplot[plot 9,bar group size={6}{10}] coordinates {(NFS,64.3)};
    \addplot[plot 12,bar group size={7}{10}] coordinates {(NFS,67.6)};
    \addplot[plot 11,bar group size={8}{10}] coordinates {(NFS,67.9)};
    \addplot[plot 13,bar group size={9}{10}] coordinates {(NFS,68.4)};
    
    \addplot[plot 3,bar group size={0}{9}] coordinates {(TNL2K,40.1)};
    \addplot[plot 4,bar group size={1}{9}] coordinates {(TNL2K,44.7)};
    \addplot[plot 5,bar group size={2}{9}] coordinates {(TNL2K,50.7)};
    \addplot[plot 17,bar group size={3}{9}] coordinates {(TNL2K,57.8)};
    \addplot[plot 8,bar group size={4}{9}] coordinates {(TNL2K,55.9)};
    \addplot[plot 9,bar group size={5}{9}] coordinates {(TNL2K,57.5)};
    \addplot[plot 12,bar group size={6}{9}] coordinates {(TNL2K,59.2)};
    \addplot[plot 11,bar group size={7}{9}] coordinates {(TNL2K,60.3)};
    \addplot[plot 13,bar group size={8}{9}] coordinates {(TNL2K,61.6)};
    \end{axis}
    \end{tikzpicture}
\caption{State-of-the-art comparison on TNL2K~\cite{wang2021tnl}, NFS~\cite{kiani2017need} and UAV123~\cite{mueller2016uav}. 
}
\label{fig:another_three}
\end{figure*}

%% file: _main.bbl
\begin{thebibliography}{10}\itemsep=-1pt

\bibitem{aiatrack}
Aiatrack: Attention in attention for transformer visual tracking.
\newblock In {\em ECCV}, pages 146--164. Springer, 2022.

\bibitem{arulampalam2002tutorial}
M~Sanjeev Arulampalam, Simon Maskell, Neil Gordon, and Tim Clapp.
\newblock A tutorial on particle filters for online nonlinear/non-gaussian bayesian tracking.
\newblock {\em IEEE Transactions on Signal Processing}, 50(2):174--188, 2002.

\bibitem{siamFC}
Luca Bertinetto, Jack Valmadre, Joao~F Henriques, Andrea Vedaldi, and Philip~HS Torr.
\newblock Fully-convolutional siamese networks for object tracking.
\newblock In {\em ECCV}, 2016.

\bibitem{bertozzi2004pedestrian}
M Bertozzi, A Broggi, A Fascioli, A Tibaldi, R Chapuis, and F Chausse.
\newblock Pedestrian localization and tracking system with kalman filtering.
\newblock In {\em IV}, pages 584--589. IEEE, 2004.

\bibitem{bhat2019learning}
Goutam Bhat, Martin Danelljan, Luc~Van Gool, and Radu Timofte.
\newblock Learning discriminative model prediction for tracking.
\newblock In {\em ICCV}, pages 6182--6191, 2019.

\bibitem{bhat2018unveiling}
Goutam Bhat, Joakim Johnander, Martin Danelljan, Fahad~Shahbaz Khan, and Michael Felsberg.
\newblock Unveiling the power of deep tracking.
\newblock In {\em ECCV}, 2018.

\bibitem{chen2022backbone}
Boyu Chen, Peixia Li, Lei Bai, Lei Qiao, Qiuhong Shen, Bo Li, Weihao Gan, Wei Wu, and Wanli Ouyang.
\newblock Backbone is all your need: A simplified architecture for visual object tracking.
\newblock In {\em ECCV}, pages 375--392. Springer, 2022.

\bibitem{chen2011kalman}
SY Chen.
\newblock Kalman filter for robot vision: a survey.
\newblock {\em IEEE Transactions on Industrial Electronics}, 59(11):4409--4420, 2011.

\bibitem{chen2021pix2seq}
Ting Chen, Saurabh Saxena, Lala Li, David~J Fleet, and Geoffrey Hinton.
\newblock Pix2seq: A language modeling framework for object detection.
\newblock {\em ICLR}, 2021.

\bibitem{chen2022unified}
Ting Chen, Saurabh Saxena, Lala Li, Tsung-Yi Lin, David~J Fleet, and Geoffrey Hinton.
\newblock A unified sequence interface for vision tasks.
\newblock In {\em NeurIPS}, 2022.

\bibitem{SeqTrack}
Xin Chen, Houwen Peng, Dong Wang, Huchuan Lu, and Han Hu.
\newblock Seqtrack: Sequence to sequence learning for visual object tracking.
\newblock In {\em CVPR}, pages 14572--14581, 2023.

\bibitem{TransT}
Xin Chen, Bin Yan, Jiawen Zhu, Dong Wang, Xiaoyun Yang, and Huchuan Lu.
\newblock Transformer tracking.
\newblock In {\em CVPR}, 2021.

\bibitem{mixformer}
Yutao Cui, Cheng Jiang, Limin Wang, and Gangshan Wu.
\newblock Mixformer: End-to-end tracking with iterative mixed attention.
\newblock In {\em CVPR}, 2022.

\bibitem{cui2023mixformerv2}
Yutao Cui, Tianhui Song, Gangshan Wu, and Limin Wang.
\newblock Mixformerv2: Efficient fully transformer tracking.
\newblock {\em NeurIPS}, 2023.

\bibitem{dai2020high}
Kenan Dai, Yunhua Zhang, Dong Wang, Jianhua Li, Huchuan Lu, and Xiaoyun Yang.
\newblock High-performance long-term tracking with meta-updater.
\newblock In {\em CVPR}, 2020.

\bibitem{danelljan2019atom}
Martin Danelljan, Goutam Bhat, Fahad~Shahbaz Khan, and Michael Felsberg.
\newblock Atom: Accurate tracking by overlap maximization.
\newblock In {\em CVPR}, 2019.

\bibitem{eco}
Martin Danelljan, Goutam Bhat, Fahad Shahbaz~Khan, and Michael Felsberg.
\newblock Eco: Efficient convolution operators for tracking.
\newblock In {\em CVPR}, 2017.

\bibitem{bert}
Jacob Devlin, Ming-Wei Chang, Kenton Lee, and Kristina Toutanova.
\newblock Bert: Pre-training of deep bidirectional transformers for language understanding.
\newblock {\em arXiv}, 2018.

\bibitem{motionaware_augment}
Shuangrui Ding, Maomao Li, Tianyu Yang, Rui Qian, Haohang Xu, Qingyi Chen, Jue Wang, and Hongkai Xiong.
\newblock Motion-aware contrastive video representation learning via foreground-background merging.
\newblock In {\em CVPR}, pages 9716--9726, 2022.

\bibitem{vit}
Alexey Dosovitskiy, Lucas Beyer, Alexander Kolesnikov, Dirk Weissenborn, Xiaohua Zhai, Thomas Unterthiner, Mostafa Dehghani, Matthias Minderer, Georg Heigold, Sylvain Gelly, Jakob Uszkoreit, and Neil Houlsby.
\newblock An image is worth 16x16 words: Transformers for image recognition at scale.
\newblock In {\em ICLR}, 2021.

\bibitem{fan2021lasot}
Heng Fan, Hexin Bai, Liting Lin, Fan Yang, Peng Chu, Ge Deng, Sijia Yu, Mingzhen Huang, Juehuan Liu, Yong Xu, Chunyuan Liao, Lin Yuan, and Haibin Ling.
\newblock Lasot: A high-quality large-scale single object tracking benchmark.
\newblock {\em International Journal of Computer Vision}, 2021.

\bibitem{fan2019lasot}
Heng Fan, Liting Lin, Fan Yang, Peng Chu, Ge Deng, Sijia Yu, Hexin Bai, Yong Xu, Chunyuan Liao, and Haibin Ling.
\newblock Lasot: A high-quality benchmark for large-scale single object tracking.
\newblock In {\em CVPR}, 2019.

\bibitem{fan2019siamese}
Heng Fan and Haibin Ling.
\newblock Siamese cascaded region proposal networks for real-time visual tracking.
\newblock In {\em CVPR}, 2019.

\bibitem{fu2021stmtrack}
Zhihong Fu, Qingjie Liu, Zehua Fu, and Yunhong Wang.
\newblock Stmtrack: Template-free visual tracking with space-time memory networks.
\newblock In {\em CVPR}, pages 13774--13783, 2021.

\bibitem{GRM}
Shenyuan Gao, Chunluan Zhou, and Jun Zhang.
\newblock Generalized relation modeling for transformer tracking.
\newblock In {\em CVPR}, pages 18686--18695, 2023.

\bibitem{gevorgyan2022siou}
Zhora Gevorgyan.
\newblock Siou loss: More powerful learning for bounding box regression.
\newblock {\em arXiv}, 2022.

\bibitem{guo2020siamcar}
Dongyan Guo, Jun Wang, Ying Cui, Zhenhua Wang, and Shengyong Chen.
\newblock Siamcar: Siamese fully convolutional classification and regression for visual tracking.
\newblock In {\em CVPR}, 2020.

\bibitem{Guo_2017_ICCV}
Qing Guo, Wei Feng, Ce Zhou, Rui Huang, Liang Wan, and Song Wang.
\newblock Learning dynamic siamese network for visual object tracking.
\newblock In {\em ICCV}, Oct 2017.

\bibitem{gustafsson2002particle}
Fredrik Gustafsson, Fredrik Gunnarsson, Niclas Bergman, Urban Forssell, Jonas Jansson, Rickard Karlsson, and P-J Nordlund.
\newblock Particle filters for positioning, navigation, and tracking.
\newblock {\em IEEE Transactions on Signal Processing}, 50(2):425--437, 2002.

\bibitem{mae}
Kaiming He, Xinlei Chen, Saining Xie, Yanghao Li, Piotr Doll{\'a}r, and Ross Girshick.
\newblock Masked autoencoders are scalable vision learners.
\newblock In {\em CVPR}, pages 16000--16009, 2022.

\bibitem{tatrack}
Kaijie He, Canlong Zhang, Sheng Xie, Zhixin Li, and Zhiwen Wang.
\newblock Target-aware tracking with long-term context attention.
\newblock {\em AAAI}, 2023.

\bibitem{huang2019got}
Lianghua Huang, Xin Zhao, and Kaiqi Huang.
\newblock Got-10k: A large high-diversity benchmark for generic object tracking in the wild.
\newblock {\em IEEE Transactions on Pattern Analysis and Machine Intelligence}, 2019.

\bibitem{javed2022visual}
Sajid Javed, Martin Danelljan, Fahad~Shahbaz Khan, Muhammad~Haris Khan, Michael Felsberg, and Jiri Matas.
\newblock Visual object tracking with discriminative filters and siamese networks: a survey and outlook.
\newblock {\em IEEE Transactions on Pattern Analysis and Machine Intelligence}, 45(5):6552--6574, 2022.

\bibitem{kiani2017need}
Hamed Kiani~Galoogahi, Ashton Fagg, Chen Huang, Deva Ramanan, and Simon Lucey.
\newblock Need for speed: A benchmark for higher frame rate object tracking.
\newblock In {\em ICCV}, 2017.

\bibitem{kiani2017learning}
Hamed Kiani~Galoogahi, Ashton Fagg, and Simon Lucey.
\newblock Learning background-aware correlation filters for visual tracking.
\newblock In {\em ICCV}, 2017.

\bibitem{slt}
Minji Kim, Seungkwan Lee, Jungseul Ok, Bohyung Han, and Minsu Cho.
\newblock Towards sequence-level training for visual tracking.
\newblock In {\em ECCV}, 2022.

\bibitem{li2019siamrpn++}
Bo Li, Wei Wu, Qiang Wang, Fangyi Zhang, Junliang Xing, and Junjie Yan.
\newblock Siamrpn++: Evolution of siamese visual tracking with very deep networks.
\newblock In {\em CVPR}, 2019.

\bibitem{li2018high}
Bo Li, Junjie Yan, Wei Wu, Zheng Zhu, and Xiaolin Hu.
\newblock High performance visual tracking with siamese region proposal network.
\newblock In {\em CVPR}, pages 8971--8980, 2018.

\bibitem{li2018learning}
Feng Li, Cheng Tian, Wangmeng Zuo, Lei Zhang, and Ming-Hsuan Yang.
\newblock Learning spatial-temporal regularized correlation filters for visual tracking.
\newblock In {\em CVPR}, 2018.

\bibitem{li2019gradnet}
Peixia Li, Boyu Chen, Wanli Ouyang, Dong Wang, Xiaoyun Yang, and Huchuan Lu.
\newblock Gradnet: Gradient-guided network for visual object tracking.
\newblock In {\em ICCV}, pages 6162--6171, 2019.

\bibitem{swin}
Liting Lin, Heng Fan, Yong Xu, and Haibin Ling.
\newblock Swintrack: A simple and strong baseline for transformer tracking.
\newblock In {\em NeurIPS}, 2022.

\bibitem{lin2014microsoft}
Tsung-Yi Lin, Michael Maire, Serge Belongie, James Hays, Pietro Perona, Deva Ramanan, Piotr Doll{\'a}r, and C~Lawrence Zitnick.
\newblock Microsoft coco: Common objects in context.
\newblock In {\em ECCV}, 2014.

\bibitem{liu2023polyformer}
Jiang Liu, Hui Ding, Zhaowei Cai, Yuting Zhang, Ravi~Kumar Satzoda, Vijay Mahadevan, and R Manmatha.
\newblock Polyformer: Referring image segmentation as sequential polygon generation.
\newblock In {\em CVPR}, pages 18653--18663, 2023.

\bibitem{weight_decay}
Ilya Loshchilov and Frank Hutter.
\newblock Decoupled weight decay regularization.
\newblock In {\em ICLR}, 2019.

\bibitem{DiMP}
Alan Luke{\v{z}}ic, Tom{\'a}{\v{s}} Voj{\'\i}r, Luka~Cehovin Zajc, Jir{\'\i} Matas, and Matej Kristan.
\newblock Discriminative correlation filter with channel and spatial reliability.
\newblock In {\em CVPR}, 2017.

\bibitem{mueller2016uav}
Matthias Mueller, Neil Smith, and Bernard Ghanem.
\newblock A benchmark and simulator for uav tracking.
\newblock In {\em ECCV}, 2016.

\bibitem{muller2018trackingnet}
Matthias Muller, Adel Bibi, Silvio Giancola, Salman Alsubaihi, and Bernard Ghanem.
\newblock Trackingnet: A large-scale dataset and benchmark for object tracking in the wild.
\newblock In {\em ECCV}, 2018.

\bibitem{nam2016learning}
Hyeonseob Nam and Bohyung Han.
\newblock Learning multi-domain convolutional neural networks for visual tracking.
\newblock In {\em CVPR}, 2016.

\bibitem{motiontrack}
Zheng Qin, Sanping Zhou, Le Wang, Jinghai Duan, Gang Hua, and Wei Tang.
\newblock Motiontrack: Learning robust short-term and long-term motions for multi-object tracking.
\newblock In {\em CVPR}, pages 17939--17948, 2023.

\bibitem{raffel2020exploring}
Colin Raffel, Noam Shazeer, Adam Roberts, Katherine Lee, Sharan Narang, Michael Matena, Yanqi Zhou, Wei Li, and Peter~J Liu.
\newblock Exploring the limits of transfer learning with a unified text-to-text transformer.
\newblock {\em JMLR}, 2020.

\bibitem{cttrack}
Zikai Song, Run Luo, Junqing Yu, Yi-Ping~Phoebe Chen, and Wei Yang.
\newblock Compact transformer tracker with correlative masked modeling.
\newblock {\em AAAI}, 2023.

\bibitem{song2022transformer}
Zikai Song, Junqing Yu, Yi-Ping~Phoebe Chen, and Wei Yang.
\newblock Transformer tracking with cyclic shifting window attention.
\newblock In {\em CVPR}, pages 8791--8800, 2022.

\bibitem{tong2022videomae}
Zhan Tong, Yibing Song, Jue Wang, and Limin Wang.
\newblock Videomae: Masked autoencoders are data-efficient learners for self-supervised video pre-training.
\newblock {\em NeurIPS}, 35:10078--10093, 2022.

\bibitem{van2008visualizing}
Laurens Van~der Maaten and Geoffrey Hinton.
\newblock Visualizing data using t-sne.
\newblock {\em Journal of Machine Learning Research}, 9(11), 2008.

\bibitem{SiamRCNN}
Paul Voigtlaender, Jonathon Luiten, Philip~HS Torr, and Bastian Leibe.
\newblock Siam r-cnn: Visual tracking by re-detection.
\newblock In {\em CVPR}, 2020.

\bibitem{wang2020tracking}
Guangting Wang, Chong Luo, Xiaoyan Sun, Zhiwei Xiong, and Wenjun Zeng.
\newblock Tracking by instance detection: A meta-learning approach.
\newblock In {\em CVPR}, pages 6288--6297, 2020.

\bibitem{videomaev2}
Limin Wang, Bingkun Huang, Zhiyu Zhao, Zhan Tong, Yinan He, Yi Wang, Yali Wang, and Yu Qiao.
\newblock Videomae v2: Scaling video masked autoencoders with dual masking.
\newblock In {\em CVPR}, pages 14549--14560, 2023.

\bibitem{TrDiMP}
Ning Wang, Wengang Zhou, Jie Wang, and Houqiang Li.
\newblock Transformer meets tracker: Exploiting temporal context for robust visual tracking.
\newblock In {\em CVPR}, 2021.

\bibitem{wang2021tnl}
Xiao Wang, Xiujun Shu, Zhipeng Zhang, Bo Jiang, Yaowei Wang, Yonghong Tian, and Feng Wu.
\newblock Towards more flexible and accurate object tracking with natural language: Algorithms and benchmark.
\newblock In {\em CVPR}, 2021.

\bibitem{ARTrack}
Xing Wei, Yifan Bai, Yongchao Zheng, Dahu Shi, and Yihong Gong.
\newblock Autoregressive visual tracking.
\newblock In {\em CVPR}, pages 9697--9706, 2023.

\bibitem{wei2023sparse}
Xing Wei, Anjia Cao, Funing Yang, and Zhiheng Ma.
\newblock Sparse parameterization for epitomic dataset distillation.
\newblock In {\em NeurIPS}, 2023.

\bibitem{weng2006video}
Shiuh-Ku Weng, Chung-Ming Kuo, and Shu-Kang Tu.
\newblock Video object tracking using adaptive kalman filter.
\newblock {\em Journal of Visual Communication and Image Representation}, 17(6):1190--1208, 2006.

\bibitem{stark}
Bin Yan, Houwen Peng, Jianlong Fu, Dong Wang, and Huchuan Lu.
\newblock Learning spatio-temporal transformer for visual tracking.
\newblock In {\em ICCV}, 2021.

\bibitem{yang2018learning}
Tianyu Yang and Antoni~B Chan.
\newblock Learning dynamic memory networks for object tracking.
\newblock In {\em ECCV}, pages 152--167, 2018.

\bibitem{yang2020roam}
Tianyu Yang, Pengfei Xu, Runbo Hu, Hua Chai, and Antoni~B Chan.
\newblock Roam: Recurrently optimizing tracking model.
\newblock In {\em CVPR}, pages 6718--6727, 2020.

\bibitem{ostrack}
Botao Ye, Hong Chang, Bingpeng Ma, and Shiguang Shan.
\newblock Joint feature learning and relation modeling for tracking: A one-stream framework.
\newblock In {\em ECCV}, 2022.

\bibitem{videomix}
Sangdoo Yun, Seong~Joon Oh, Byeongho Heo, Dongyoon Han, and Jinhyung Kim.
\newblock Videomix: Rethinking data augmentation for video classification.
\newblock {\em arXiv}, 2020.

\bibitem{motr}
Fangao Zeng, Bin Dong, Yuang Zhang, Tiancai Wang, Xiangyu Zhang, and Yichen Wei.
\newblock Motr: End-to-end multiple-object tracking with transformer.
\newblock In {\em ECCV}, pages 659--675. Springer, 2022.

\bibitem{zhang2019learning}
Lichao Zhang, Abel Gonzalez-Garcia, Joost Van~De Weijer, Martin Danelljan, and Fahad~Shahbaz Khan.
\newblock Learning the model update for siamese trackers.
\newblock In {\em ICCV}, pages 4010--4019, 2019.

\bibitem{Zhang_2019_ICCV}
Lichao Zhang, Abel Gonzalez-Garcia, Joost van~de Weijer, Martin Danelljan, and Fahad~Shahbaz Khan.
\newblock Learning the model update for siamese trackers.
\newblock In {\em ICCV}, October 2019.

\bibitem{motrv2}
Yuang Zhang, Tiancai Wang, and Xiangyu Zhang.
\newblock Motrv2: Bootstrapping end-to-end multi-object tracking by pretrained object detectors.
\newblock In {\em CVPR}, pages 22056--22065, 2023.

\bibitem{Ocean}
Zhipeng Zhang, Houwen Peng, Jianlong Fu, Bing Li, and Weiming Hu.
\newblock Ocean: Object-aware anchor-free tracking.
\newblock In {\em ECCV}, 2020.

\bibitem{zhu2018distractor}
Zheng Zhu, Qiang Wang, Bo Li, Wei Wu, Junjie Yan, and Weiming Hu.
\newblock Distractor-aware siamese networks for visual object tracking.
\newblock In {\em ECCV}, 2018.

\end{thebibliography}
